\documentclass[lettersize,journal]{IEEEtran}
\usepackage{amsmath,amssymb,amsfonts}
\usepackage{algorithmic}
\usepackage{algorithm}
\usepackage{array}
\usepackage[caption=false,font=normalsize,labelfont=rm,textfont=rm]{subfig}
\usepackage{textcomp}
\usepackage{stfloats}
\usepackage{url}
\usepackage{verbatim}
\usepackage{graphicx}
\usepackage{cite}
\usepackage{xcolor}

\definecolor{myPurple}{RGB}{123,17,158}

\usepackage{booktabs}
\usepackage{multirow}
\usepackage{makecell}
\usepackage{copyrightbox}
\usepackage{threeparttable}
\usepackage[linkcolor=myPurple,citecolor=myPurple,urlcolor=blue,colorlinks=true]{hyperref}
\usepackage{hyperref}
\usepackage{cleveref}
\usepackage{caption}
\usepackage{colortbl}      
\usepackage{pifont}

\newcommand{\cmark}{\textcolor{green}{\ding{51}}} 
\newcommand{\xmark}{\textcolor{red}{\ding{55}}}   
\newcommand{\etal}{\textit{et al.}}
\definecolor{mygreen}{HTML}{32C016}

\begin{document}
\title{ToFormer: Towards Large-scale Scenario Depth Completion for Lightweight ToF Camera}
\author{Juncheng Chen$^{1,2}$, Tiancheng Lai$^{1,2}$, Xingpeng Wang$^{1,2}$, Bingxin Liao$^{2}$, Baozhe Zhang$^{2,3}$, \\ Chao Xu$^{1,2}$, and Yanjun Cao$^{1,2,*}$ 
\thanks{$^{1}$State Key Laboratory of Industrial Control Technology, Institute of Cyber Systems and Control, Zhejiang University, Hangzhou, China.}
\thanks{$^{2}$Huzhou Institute of Zhejiang University, Huzhou, China.}
\thanks{$^{3}$The Chinese University of Hong Kong, Shenzhen, China.}
\thanks{$^{*}$Corresponding author. This work was supported by China National Tobacco Corporation's Key R\&D Program (Grant No. 110202402018).}}
\markboth{ }%
{ToFormer: Towards Large-scale Scenario Depth Completion for Lightweight Time-of-Flight Camera}

\maketitle

\begin{abstract}
Time-of-Flight (ToF) cameras possess compact design and high measurement precision to be applied to various robot tasks. However, their limited sensing range restricts deployment in large-scale scenarios. Depth completion has emerged as a potential solution to expand the sensing range of ToF cameras, but existing research lacks dedicated datasets and struggles to generalize to ToF measurements. In this paper, we propose a full-stack framework that enables depth completion in large-scale scenarios for short-range ToF cameras. First, we construct a multi-sensor platform with a reconstruction-based pipeline to collect real-world ToF samples with dense large-scale ground truth, yielding the first \textbf{LA}rge-\textbf{S}cal\textbf{E} scena\textbf{R}io ToF depth completion dataset (LASER-ToF). Second, we propose a sensor-aware depth completion network that incorporates a novel 3D branch with a 3D-2D Joint Propagation Pooling (JPP) module and Multimodal Cross-Covariance Attention (MXCA), enabling effective modeling of long-range relationships and efficient 3D-2D fusion under non-uniform ToF depth sparsity. Moreover, our network can utilize the sparse point cloud from visual SLAM as a supplement to ToF depth to further improve prediction accuracy. Experiments show that our method achieves an 8.6\% lower mean absolute error than the second-best method, while maintaining lightweight design to support onboard deployment. Finally, to verify the system's applicability on real robots, we deploy proposed method on a quadrotor at a 10\,Hz runtime, enabling reliable large-scale mapping and long-range planning in challenging environments for short-range ToF cameras.
 
\def\abstractname{Note to Practitioners}
\begin{abstract}
Lightweight robots can be equipped with ToF cameras, which provide accurate but short-range depth and therefore struggle in large scenarios such as outdoor fields, warehouses, substations, or factories. This work offers a practical solution for extending the sensing range of ToF cameras with edge-computing applicability. We develop a toolchain that allows practitioners to collect depth completion datasets for their own ToF cameras, together with a depth completion network that turns short-range ToF and RGB image into dense and long-range depth. The system can optionally incorporate visual-SLAM cues for improved robustness in practical deployments. We demonstrate that the network runs in real time on a small quadrotor, enabling it to perform large-scale dense mapping and to plan safer and more efficient paths. All hardware designs, software tools, trained models, and datasets will be open-sourced so that practitioners can directly adapt them to their applications. One current limitation is the need for careful RGB-ToF cameras calibration. Future work will focus on tighter integration with SLAM and release more potential applications of ToF depth completion. 
\end{abstract}


\end{abstract} 
\section{Introduction}
\label{sec:intro}

\IEEEPARstart{T}{ime-of-Flight} (ToF) cameras have become increasingly attractive
for robotic perception due to their accurate depth measurement, compact size,
and low power consumption. Compared to stereo cameras, ToF cameras typically provide more stable depth measurements with fewer texture dependencies, while being significantly lighter and more energy-efficient than LiDAR sensors. These properties make ToF cameras well suited for robot applications~\cite{kinectfusion,torf,6DoF,ToF-SLAM}. For instance, T{\"o}RF~\cite{torf} exploits accurate ToF measurements as priors for dynamic scene reconstruction, while Hochdorfer \etal~\cite{6DoF} and Chen \etal~\cite{ToF-SLAM} leverage the infrared characteristics of ToF cameras to enable robust visual SLAM (Simultaneous Localization and Mapping) in low-light environments. Despite these successes, the limited sensing range of ToF cameras remains a critical bottleneck, largely confining their application to small-scale indoor scenarios. Extending the effective sensing range of ToF cameras through depth completion has therefore emerged as a promising direction. However, existing studies on ToF depth completion still exhibit several fundamental challenges:

\textbf{(1) Lack of large-scale scene dataset:} Community lacks a dataset or benchmark specifically designed for ToF camera depth completion in large-scale scenes. As shown in Table~\ref{dataset}, RGB-D datasets like NYU-Depth V2 \cite{NYUv2} and TOFDC \cite{TOFDC} lack depth supervision for pixels beyond 6\,m. LiDAR datasets like KittiDC \cite{KittiDC} are designed for spacious outdoor but do not include ToF data or dense per-pixel ground truth. Consequently, obtaining a large-scale, high-density, and consecutive-frame depth completion dataset for real-world ToF sampling remains a non-trivial task yet to be undertaken.

\begin{figure*}[th]
  \captionsetup{font={small}}
  \centerline{\includegraphics[width=\textwidth]{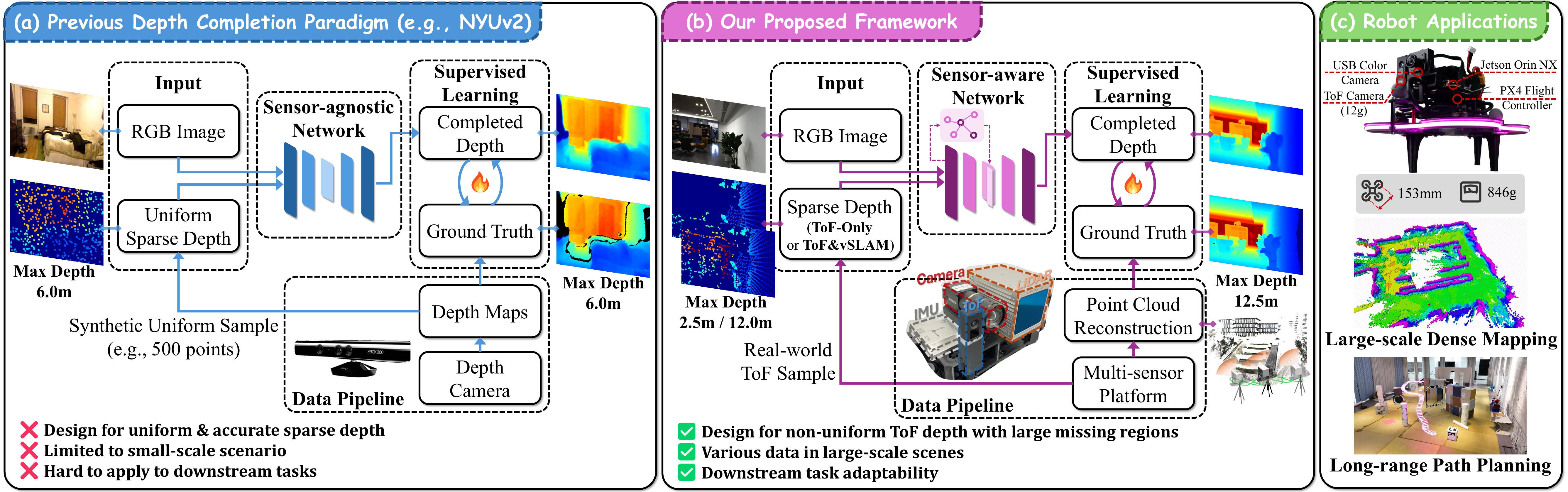}}
  \caption{\textbf{(a) Previous paradigms} typically acquire ground truth through a depth camera and subsample it to synthesis uniform sparse inputs for training. Sensor-agnostic networks developed under this paradigm struggle to transfer to real-world ToF completion. \textbf{(b) Our framework} first focuses on obtaining real-world ToF sampling and generating dense depth ground truth in large-scale scenarios. Then, we propose a sensor-aware depth completion network. This framework expands the applicability of ToF cameras in large-scale robot applications in \textbf{(c)}.}
  \label{title_fig}
\end{figure*}

\textbf{(2) Large missing regions and non-uniform depth sampling:} Due to the imaging principles, phase ambiguity, and power constraints of lightweight ToF cameras, their sensing range is typically limited to 3–6\,m (e.g., PMD Flexx2 ToF camera has a sensing range of 3\,m). This results in large missing regions in ToF depth maps. Moreover, unlike prior depth completion assumptions, where sparse depth inputs are synthetically and uniformly sampled (Fig.~\ref{title_fig} (a)), ToF depth exhibits non-uniform spatial distributions due to surface materials and physical sensing principles. As a result, existing sensor-agnostic depth completion networks~\cite{NLSPN,Dyspn,CompletionFormer,DFU}, which do not explicitly model the large missing regions and non-uniform sparsity inherent to ToF depth, face fundamental limitations.

In this work, we seek to address these challenges and promote large-scale ToF depth completion via a systematic framework (Fig.~\ref{title_fig} (b)) as well as a real-world robot validation (Fig.~\ref{title_fig} (c)). This framework enables ToF cameras to serve as lightweight depth sensors with long-range coverage, broadening their applicability in large-scale robotic tasks such as large-scale dense mapping and long-range path planning.

\textbf{First,} we create \textbf{LASER-ToF} depth, the first \textbf{LA}rge-\textbf{S}cal\textbf{E} Scena\textbf{R}io ToF depth completion dataset and benchmark, which is based on a multi-sensor platform and a LiDAR-Visual-Inertial (LVI) SLAM system. We propose a reconstruction-based method to produce accurate, per-frame, and large-scale ground truth for real-world ToF depth map. The proposed pipeline allows efficient data collection, where only 1-2 minutes of scanning are sufficient to acquire 300--500 viewpoints within a scene, while achieving an average return density of up to 94.6\%.

\textbf{Second,} we propose a sensor-aware completion network, which is lightweight yet effective for handling large missing regions and non-uniform ToF depth. The network utilizes a vision transformer (ViT) encoder to capture long-range relationship, and a novel 3D branch with 3D–2D Joint Propagation Pooling (JPP) module for dynamic non-local point cloud aggregation and dense cross-modal interaction. Then, features from RGB, depth, and 3D branch are efficiently fused via our Multimodal Cross-Covariance Attention (MXCA). Additionally, our framework also supports point clouds from visual SLAM as additional inputs to improve the overall performance of the depth completion network. Extensive experiments on the proposed LASER-ToF benchmark demonstrate that our method outperforms the second-best method by 8.6\% in mean absolute error, while reducing parameter count and runtime relative to the average baseline by 85.9\% and 73.8\%.

\textbf{Third,} we deploy the proposed network in real time on a quadrotor platform
to validate its practical applicability. With onboard inference at 10\,Hz, the quadrotor is able to reconstruct a large-scale scene of 50\,m$\times$50\,m, and reduces the energy cost and travel time in a complex autonomous planning task by 29.0\% and 16.2\%, respectively.

To summarize, our main contributions are threefold:
\begin{itemize}
    \item We construct LASER-ToF, the first real-world dataset and benchmark for large-scale ToF depth completion. 
    \item We propose a sensor-aware depth completion network, which explicitly models non-uniform ToF sampling patterns and performs multimodal fusion. 
    \item We integrate ToF depth completion into a quadrotor to validate its real-time performance and effectiveness in downstream mapping and planning tasks. 
\end{itemize}

\section{Related Work}
\label{sec:Related Work}
Depth completion aims to densify the sparse depth maps produced by depth sensors or SLAM systems, enabling complete geometric perception for downstream robotic and vision tasks. In this section, we summarize four most relevant topics, including depth completion dataset and benchmark, supervised depth completion, generalizable depth completion, and ToF depth completion.

\subsection{Depth Completion Dataset and Benchmark.} 
Datasets and benchmarks are the foundation of learning-based methods to define the sensors, sparsity patterns, and evaluation protocols. Table~\ref{dataset} presents the current popular depth completion datasets. NYUv2~\cite{NYUv2} is a widely used indoor benchmark for depth completion, providing ground truth depth up to 6\,m range from a Kinect camera. Following common practice, the ground truth is uniformly subsampled (e.g., 500 points) to create sparse inputs. TOFDC~\cite{TOFDC} employs industrial-level ToF cameras for supervision to complete ToF depth on mobile phones. TOFDC offers a way to collect annotations for ToF depth completion and mainly targets completing the small holes or edge defects. In contrast, KittiDC~\cite{KittiDC} is a LiDAR-based long-range outdoor benchmark collected with a 64-line Velodyne LiDAR, where single-frame LiDAR maps serve as sparse inputs and multi-frame accumulated maps as ground truth. Therefore, ToF depth completion falls outside the scope of KittiDC. In summary, existing dataset and benchmarks have failed to meet the two core requirements for ToF depth completion in large-scale scenarios, namely real-world ToF sampling and corresponding per-frame large-scale depth ground truth, limiting further study.

\begin{table*}[t]
    \begin{center}
    \captionsetup{font={small}}
    \caption{Statistics of LASER-ToF compared to other depth completion datasets.}
    \resizebox{1.0\textwidth}{!}{
        \begin{tabular}{l|c|c|c|c|c|r|c|c}
            \toprule
            Dataset                     & Scene    &  \makecell[c]{Ground Truth\\Acquisition Method} & \makecell[c]{Max Reliable\\ Range} & Sequence? & ToF data? & \makecell[c]{Avg. Return \\ Density} & Resolution & Quantity \\
            \midrule
            NYU-Depth V2\cite{NYUv2}    & Indoor    & \multirow{3}{*}{Depth-Camera-Based} & \multirow{2}{*}{Near, \textless6m}   & \cmark  & \xmark  & 68\%          & 304*228 & 48,238  \\
            TOFDC\cite{TOFDC}           & In/Outdoor&                                    &                                        & \xmark & \cmark  & 98.5\%        & 512*384 & 10,560 \\ \cline{4-4} \noalign{\vskip 0.5mm}
            VOID\cite{VOID}             & In/Outdoor&                                    & Near, \textless3m                      & \cmark & \xmark  & 95.1\%        & 640*480 & 40,800  \\ 
            \midrule
            DenseLivox\cite{DenseLivox} & In/Outdoor& \multirow{3}{*}{Accumulation-Based} & \multirow{4}{*}{Far, \textgreater20m}    & \xmark & \xmark  & 88.3\%       & Unknown  & 19,428 \\
            DIODE\cite{DIODE}           & In/Outdoor&                                      &                                       & \xmark  & \xmark & 99.6\%/66.9\% & 1024*768 & 27,858  \\
            KittiDC\cite{KittiDC}       & Outdoor   &                                      &                                      & \cmark & \xmark & 22\%          & 1216*256 & 87,898 \\ \cline{3-3} \noalign{\vskip 0.5mm}
            LASER-ToF (Ours)            & In/Outdoor& Reconstruction-Based                 &                                      & \cmark & \cmark & 94.6\%        & 640*480  & 20,996  \\ 
            \bottomrule
        \end{tabular}
    } 
    
    \label{dataset}
    \end{center}
\end{table*}

\subsection{Supervised Depth Completion.} 
Supervised learning is mainstream paradigm and offers diverse entry points. Multi-modal fusion is often the first to be considered \cite{FusionNet,PENet,MDANet,CNNT,CompletionFormer}, which bridges the gap between RGB features and depth features. Some of these methods, such as CFormer~\cite{CompletionFormer} and PENet~\cite{PENet}, use additional spatial propagation network (SPN) stage to refine the final depth prediction. Specifically, SPN-based models \cite{cspn,cspn++,NLSPN,Dyspn} focus on learning an affinity matrix to propagate existing depth information to neighbor pixels. Considering that direct 2D feature extraction for RGBD images learns 3D geometric relationships in an implicit yet ineffective manner \cite{DC_survey}, some studies have started to explore explicit 3D representation for sparse or semi-dense depth map \cite{PointFusion,GAENet, CostDCNet,PointDC,TOFDC}. PointFusion~\cite{PointFusion} is an early work that explores multi-scale 3D point fusion with image features. Recently, LRRU~\cite{LRRU}, DFU~\cite{DFU}, and BP-Net~\cite{BP-Net} are dedicated to more effective multi-levels propagation. However, previous methods generally ask for uniformly sampled sparse depth input without fitting real-world requirements, such as non-uniform pattern and large missing regions in ToF depth maps.  

\subsection{Generalizable Depth Completion.} 
Recently, multiple generalizable methods emerge and gradually transcend traditional supervised methods under cross-domain settings. Most generalizable methods either rely on large foundation models~\cite{viola2025marigold, wang2025pacgdc}, or are trained extensively across diverse datasets~\cite{zuo2025omni}. Afterward, they are transferred in a zero-shot manner to new test domains with different depth sampling patterns (e.g., uniformly sampled depth, COLMAP~\cite{Colmap} reconstruction, visual SLAM maps, and LiDAR depth). 
Marigold-DC~\cite{viola2025marigold} builds on a pretrained diffusion-based depth model and is further fine-tuned on synthetic depth completion samples. PacGDC~\cite{wang2025pacgdc} synthesizes pseudo depth labels using multiple depth foundation models to enrich training diversity. OMNI-DC~\cite{zuo2025omni} leverages a mixture of high-quality datasets together with scale normalization and synthetic sparse depth patterns. Despite their strong generalization ability, these approaches predominantly use synthetic sparse depth patterns and pay limited attention to ToF cameras or scenarios where the scene scale is significantly larger than the effective range of depth sensors. More importantly, these methods incur substantial computational costs, making it impractical to deploy them to edge devices.

\subsection{ToF Depth Completion.} 
Recent works have explored ToF depth completion. Wild ToFu~\cite{WildToF} directly utilizes raw correlation images from ToF cameras as input and uses RealSense D435 for supervision. Jiang \etal~\cite{tof-low} and SpAgNet~\cite{SADC} sample from the ground truth of NYUv2~\cite{NYUv2} dataset to acquire simulated ToF depth, which differs from real-world ToF sampling. TOFDC~\cite{TOFDC} mainly targets completion for small holes or edge defects without considering challenges in large-scale scenarios. Due to the simulated assumption of ToF depth pattern and limited range of ground truth depth, above methods are difficult to directly transfer to real-world applications, particularly when the scene scale increases and large missing regions arise in ToF depth maps.

\section{LASER-ToF Depth Dataset}
\label{sec:dataset}

LASER-ToF depth dataset contains 52 sequences in large-scale scenes for ToF depth completion, totaling 20,996 frames. There are 35 indoor sequences and 17 outdoor sequences. The average scene depth range extends to 26.3\,m, which is far beyond the typical operational range of commodity ToF cameras (roughly 3–6\,m). The depth ground truth has an average return density of 94.6\% (\textit{i.e.}, the ratio of color pixels with depth measurements to all color pixels). We named this dataset the \textbf{LA}rge-\textbf{S}cal\textbf{E} Scena\textbf{R}io dataset for ToF depth completion, abbreviated as \textbf{LASER-ToF} depth. 

\begin{figure*}[th]
    \captionsetup{font={small}}
    \centerline{\includegraphics[width=1.0\textwidth]{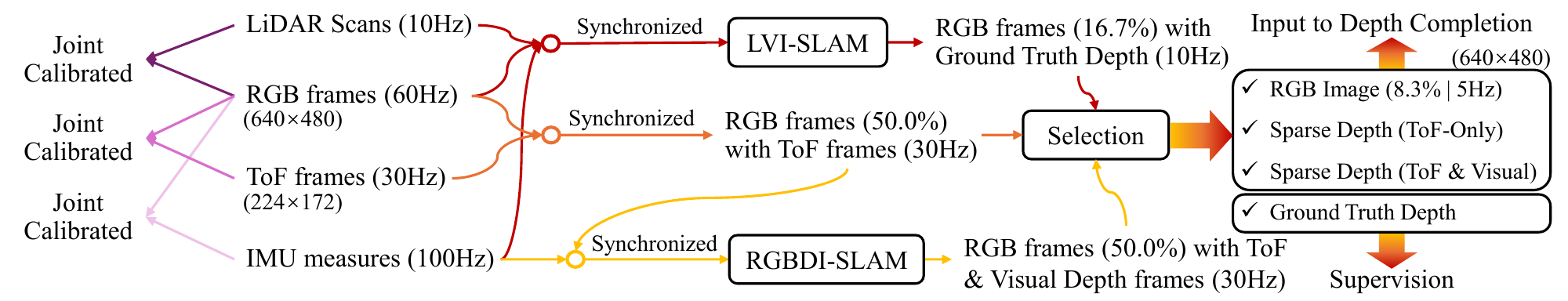}}
    \caption{\textbf{Data synchronization and dataset production pipeline.}}
    \label{sync_data}
    \vspace{-10pt}
\end{figure*}

\begin{figure}[t]
    \captionsetup{font={small}}
    \centerline{\includegraphics[width=0.48\textwidth]{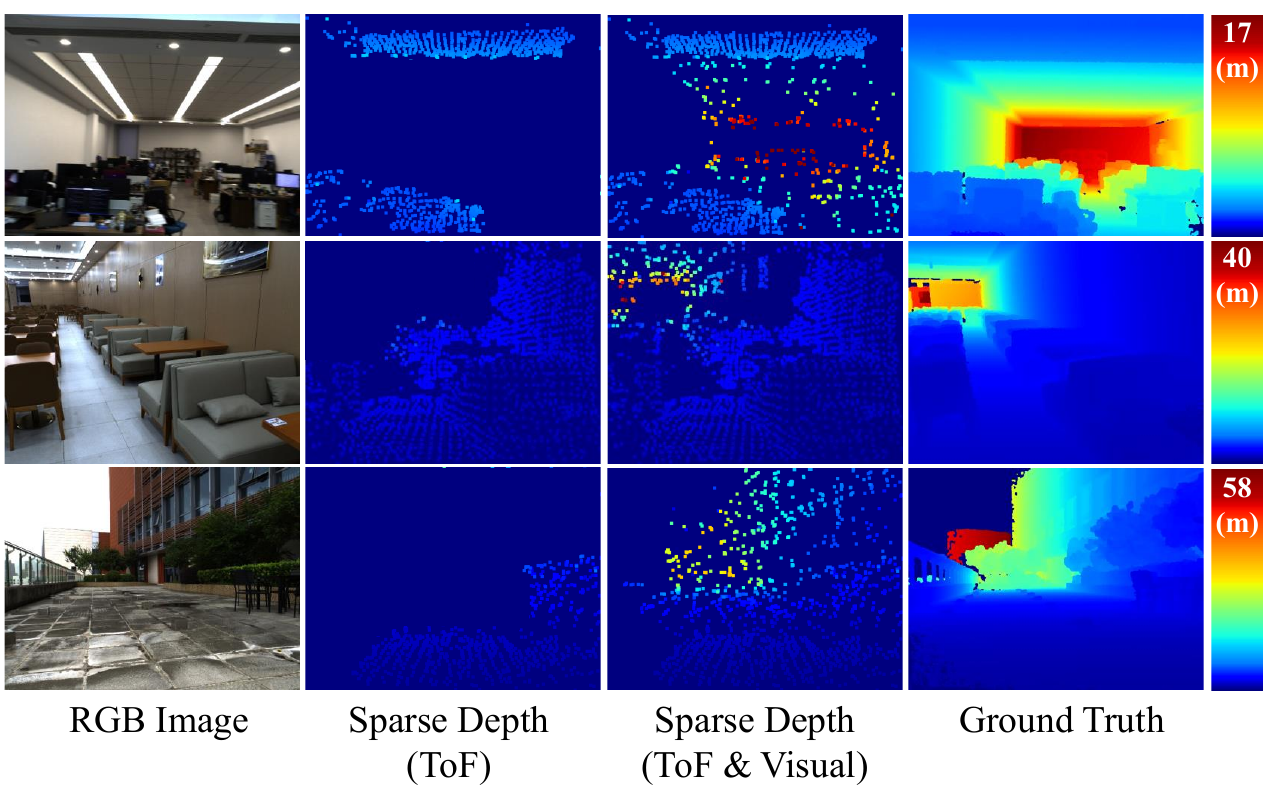}}
    \caption{\textbf{Dataset visualization: RGB image (input), sparse depth (input), and ground truth.} Color bars on the right indicate the farthest distance of depth maps.}
    \label{datavis}
\end{figure}

Table~\ref{dataset} compares the statistics between existing depth completion datasets and LASER-ToF. Compared to existing depth-camera-based datasets, LASER-ToF offers a significantly larger range of available depth. In contrast to accumulation-based datasets collected by LiDAR, LASER-ToF possesses consecutive sequence sampling and higher return density while including ToF data. Dataset samples are visualized in Fig.~\ref{datavis}.

\subsection{Multi-sensor Data Collection Platform}
\begin{figure}[th]
    \captionsetup{font={small}}
    \centerline{\includegraphics[width=0.48\textwidth]{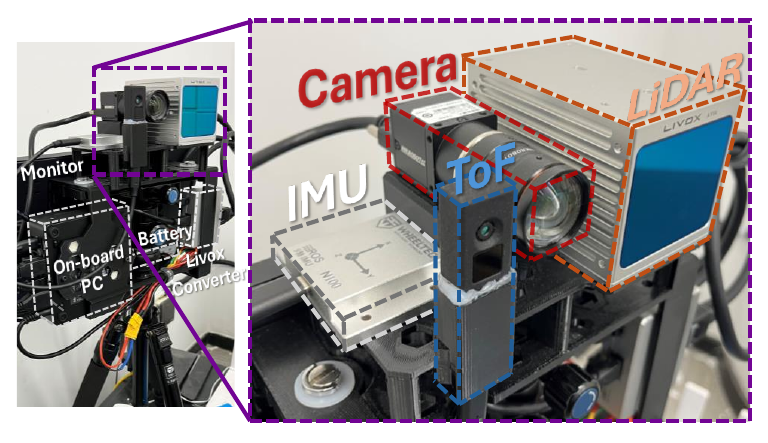}}
    \caption{\textbf{Our multi-sensor data collection platform.}}
    \label{Sensor_suppl}
\end{figure}

To collect and build LASER-ToF, we construct a multi-sensor platform with a
LiDAR-Visual-ToF-Inertial setup, which is shown in Fig.~\ref{Sensor_suppl} and includes:
\begin{itemize}
    \item Livox Avia, a solid-state LiDAR, with a field of view (FoV) of 70.4°$\times$77.2° (H$\times$V) and point rate of 240,000 points/s. It provides high-density and high-precision scans, with a range precision of 2\,cm at 20\,m and a detection range of 190\,m for surfaces with 10\% reflectivity.
    \item HIKROBOT MV-CS020-10UC, a global shutter camera with a FoV of 62°$\times$44° (H$\times$V) and a resolution of 640$\times$480. 
    \item PMD Flexx2, a lightweight ToF camera, with a depth resolution of 224$\times$172 and a FoV of 56°$\times$44°, provides depth maps within 3\,m (depends on the surface material) at 30\,Hz.
    \item Wheeltec N100, a consumer-level inertial measurement unit (IMU).
\end{itemize}

Fig.~\ref{sync_data} shows our data synchronization and dataset production pipeline. As a premise, RGB camera is jointly calibrated with LiDAR, ToF, and IMU respectively through approaches provided by Yuan \etal~\cite{MARS-Calib} and Kalibr~\cite{kalibr}. To synchronize data frames, we first soft-synchronize RGB (60\,Hz) and LiDAR frames (10\,Hz), ensuring that 16.7\% (10\,Hz / 60\,Hz) of the RGB frames align with LiDAR scans. These RGB frames can obtain corresponding dense ground truth depth maps in LVI SLAM. Meanwhile, we soft-synchronize RGB (60\,Hz) and ToF frames (30\,Hz) so that 50\% (30\,Hz / 60\,Hz) of the RGB frames correspond to ToF frames. With synchronized RGB-ToF-Inertial frames, we run RGBD-I SLAM to obtain sparse depth from visual point clouds. Finally, we select the RGB frames which possess corresponding sparse depth frames and ground truth depth. These aligned RGB, Ground Truth, and Sparse Depth triplets are generated at a frequency of 5\,Hz  (50\% of RGB-Ground Truth frames at 10\,Hz) and serve as training data.

Through this platform and pipeline, we can obtain dense ground truth depth, RGB image, and two types of sparse depth, including raw ToF depth (ToF Only) and ToF\&Visual depth.

\subsection{Dense Depth Ground Truth}\label{subsec:gt}
Acquiring dense depth ground truth for scenes has always been a challenge \cite{DC_survey,DC_survey2}. 
Previous methods generally employ statically accumulating LiDAR scans to obtain dense long-range ground truth. For instance, DIODE \cite{DIODE} requires the scanner to remain stationary for 11 minutes to capture panoramic depth ground truth. Accumulation-based methods \cite{DenseLivox,DIODE} usually fail to provide consecutive ground truth sequences, while depth-camera-based methods generate consecutive but short-range ground truth. We aim to obtain dense supervision for each frame in ToF depth sequence, thus proposing a reconstruction-based method utilizing LVI-SLAM and 3D reconstruction. 

Our ground truth acquisition is based on R\textsuperscript{3}LIVE \cite{r3live}, a LiDAR-Visual-Inertial tightly-coupled state estimation and mapping system, which is able to reconstruct precise, dense surrounding environment and provide accurate sensor poses in real time. Given the point cloud map ${M_{LVI}}$ and the current camera pose ${[R|t]}$, we first perform projection to obtain an initial depth map.
Then, we apply a minimum filter for occlusion detection to remove invisible background points and obtain depth ground truth ${D_{gt}}$ as follow:
\begin{equation}
    MinFilter(I) = \min_{(i, j) \in \mathcal{W}} I(u+i, v+j),  \forall (u, v) \in I ,
\end{equation}
\begin{equation}
    D_{gt} = MinFilter(Proj_{K}^{[R|t]}(M_{LVI})) ,
\end{equation} 
where $I$ represents the input image, $\mathcal{W}$ represents the sliding window (e.g., 3$\times$3, 5$\times$5), and $K$ represents camera's intrinsic parameter. 

Notably, previous accumulation-based methods can be affected by the FoV mismatch between camera and LiDAR, which results in missing depth values for certain image regions. For example, the upper part of images in KittiDC~\cite{KittiDC} contains no valid depth returns. In contrast, our approach reconstructs the entire scene and retrieves depth for each pixel in the camera coordinate system by directly querying the reconstructed scene point cloud. As a result, our ground truth acquisition is not constrained by camera–LiDAR FoV discrepancies. 

\subsection{Depth from Visual Feature Points}\label{subsec:vslam}

Lightweight robots (such as quadcopters) often acquire localization information from visual SLAM~\cite{zhou2020ego, zhou2021fuel}. The position of visual point clouds in SLAM systems~\cite{Vins, ORB-SLAM3} highly depends on environment and visual keypoint type, such as Oriented FAST \cite{orb_feature}, Shi-Tomasi \cite{gftt}, and Superpoint \cite{superpoint}. Although the arbitrary and unconstrained sampling of visual point clouds leads to non-uniform spatial distributions, along with noise introduced by triangulation, they scatter in those distant regions where ToF depth is missing (see Fig.~\ref{datavis}). These points can be utilized by depth completion models if visual SLAM system is available~\cite{PointFusion,VOID}. Our dataset additionally provides these visual point cloud depths as an optional form of sparse depth. 

Specifically, we take the synchronized RGB images, ToF depth maps, and IMU data as inputs to the visual SLAM system. We employ the highly accurate and widely adopted ORB-SLAM3~\cite{ORB-SLAM3} to ensure stable and continuous tracking during data acquisition. The local visual point cloud map maintained by the SLAM system is then projected onto the ToF-only sparse depth maps to generate the ToF-Visual sparse depth maps.
\section{Network Architecture}
\label{sec:model}

The primary challenge to completing ToF depth in large-scale scenarios is the sparse depth input with non-uniformity and large missing regions, which are visualized in Fig.~\ref{datavis}. Furthermore, when visual point clouds are used as additional sparse depth inputs, they still exhibit non-uniform spatial distributions and remain far from forming dense depth maps, while also introducing observation noise and outliers from SLAM tracking.

\begin{figure*}[thbp]
    \captionsetup{font={small}}
    \centerline{\includegraphics[width=1.0\textwidth]{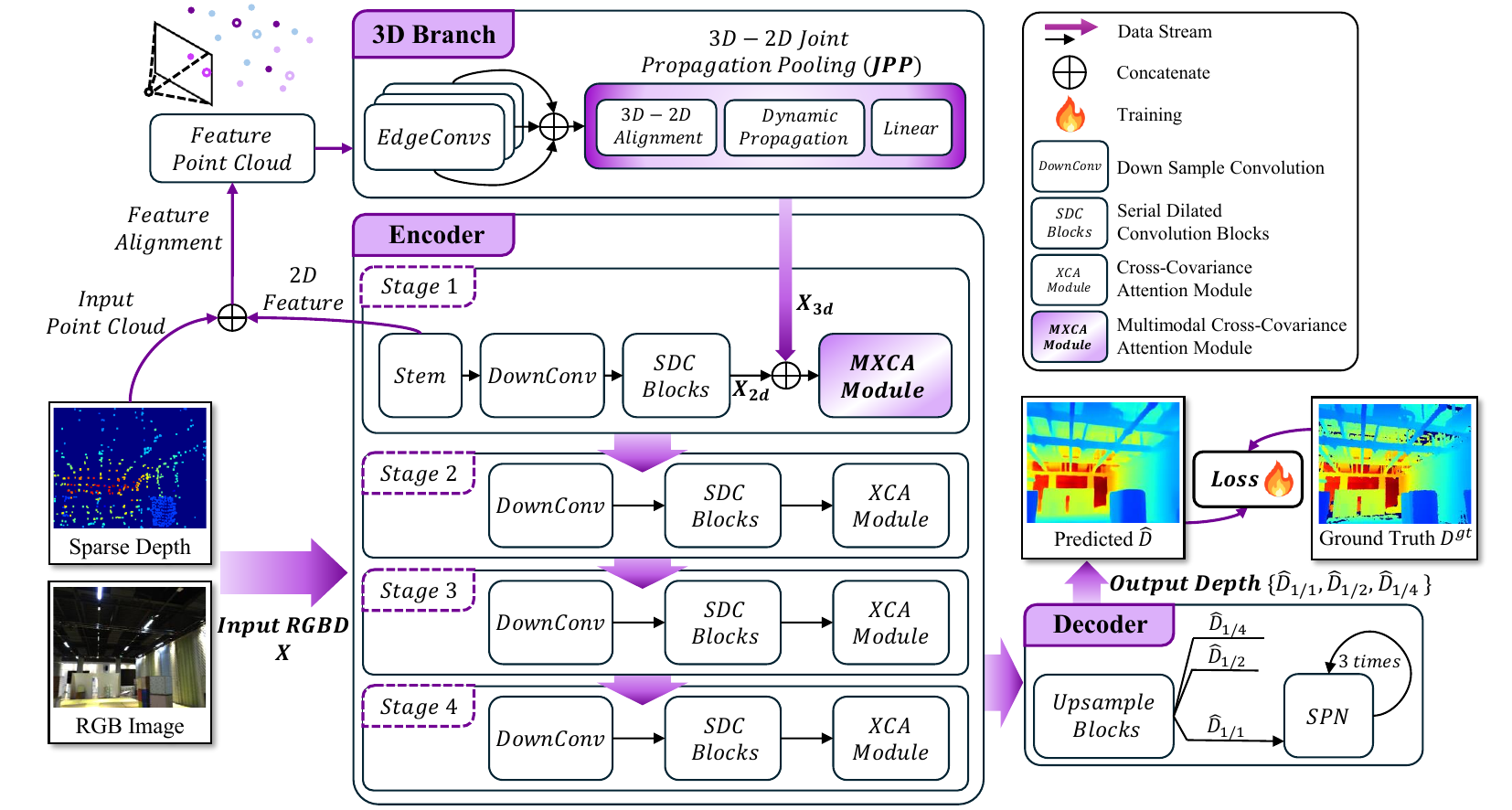}}
    \caption{\textbf{ToFormer Network Architecture.} Given an input RGBD image, a multi-scale encoder is adopted for 2D RGB-D fusion and 3D-2D cross-modal fusion. Point clouds are backprojected from RGBD image, aligned with 2D features, and fed into the 3D branch. Early fusion is conducted between the 3D feature map $X_{3d}$ and 2D feature map $X_{2d}$ through the MXCA module. Then, multi-level depth maps are upsampled by the decoder, and the full-resolution depth map $\hat{D}_{1/1}$ will be refined by the SPN module.}
    \label{Network}
\end{figure*}

To handle non-uniform depth pattern and large missing regions in large-scale ToF depth completion, our sensor-aware depth completion network (shown in Fig.~\ref{Network}) integrates three modules:
(1) \textbf{Encoder} (Sec.~\ref{sec:encoder}), to capture long-range appearance relationships between depth-available and depth-missing regions through 2D RGB-D fusion and 3D-2D cross-modal fusion. 
(2) \textbf{3D branch} (Sec.~\ref{sec:3d_branch}), to model relative geometric relationships of point cloud and enable efficient dense 3D-2D fusion. 
(3) \textbf{Decoder} (Sec.~\ref{sec:decoder}), to reconstruct final depth prediction through multi-scale upsampling and dynamic SPN refinement.

Our network takes an RGB-D image $X \in \mathbb{R}^{H \times W \times 4}$ as input, formed by concatenating an RGB image and a sparse depth map where missing values are zero-filled, and outputs a complete depth prediction $\hat{D}$. During training, the network is supervised by the depth ground truth $D^{gt}$ through loss function (Sec.~\ref{sec:loss}). 

\subsection{Encoder}\label{sec:encoder}

\subsubsection{2D RGB-D Fusion}
We designed a CNN-Transformer hybrid encoder to perform 2D RGB-D fusion (Fig.~\ref{Network} encoder part). We adopt serial dilated convolution (SDC) to enlarge receptive field for local 2D features. We leverage cross-covariance attention (XCA)~\cite{xcit} to model long-range 2D appearance relationships while maintaining linear computational complexity with respect to image size. We use down sample convolution (DownConv) to halve the spatial resolution of feature maps. The encoder contains four sequential stages, where the output of each stage serves as the input to the next. Each stage consists of DownConv, SDC blocks, and XCA module. Except for stage~1, it additionally includes a stem module for pre-feature extraction and an multimodal cross-covariance attention (MXCA) module for 3D-2D fusion.

At the beginning of each stage, the input feature maps are down sampled by DownConv and sent to SDC blocks to capture local features. The SDC blocks include a series of depth-wise dilated convolutions, batch normalization layers, and GELU \cite{GELU} activation layers. Following the suggestions of \cite{wang2018understanding}, we configure the SDC blocks with incremental dilation rates to ensure a fine and efficient receptive field.

After the SDC blocks, we employ XCA to perform global feature interaction as following formulation: 
\begin{equation}
    XCA_{Attention}(Q,K,V)=V\cdot Softmax(Q^\top K/\tau), 
\end{equation}
where $Q=XW_q$, $K=XW_k$, $V=XW_v$, and $X,Q,K,V\in \mathbb{R}^{N\times d}$.
For $X$ as the input matrix, $Q$, $K$, and $V$ represent the queries, keys, and values obtained by linearly projecting $X$, where $W_q$, $W_k$, and $W_v$ denote learnable linear projection matrices. An adaptive scaling parameter $\tau$ is used to adjust the distribution of attention weights. 

Note that XCA attention is computed along the dimensionality $d$ rather than the number of tokens $N$. Let $h$ be the number of attention heads, the time complexity of this attention mechanism is $O(N*d^2/h)$, which scales linearly with the number of tokens. In contrast, the original attention mechanism~\cite{attention} consumes a computational cost of $O(N^2*d)$, which increases quadratically with the number of tokens. Together with the use of depth-wise dilated convolutions in the SDC blocks, the XCA ensures lightweight design of the encoder.

\subsubsection{3D-2D Cross-Modal Fusion}
We introduce a variant of cross-covariance attention (XCA), \textbf{multimodal cross-covariance attention} (MXCA), which bridges the cross-modal representation gap and performs 3D-2D cross-modal fusion at stage~1 of encoder. Let $X_{2d}$ represent 2D feature map extracted from preceding steps in stage~1, $X_{3d}$ denotes 3D feature map from 3D branch (will be elaborated in the next subsection). $\bar{X}_{2d}\in \mathbb{R}^{N\times d_{2d}}$ and $\bar{X}_{3d}\in \mathbb{R}^{N\times d_{3d}}$ are tokenized 2D and 3D feature maps. We first concatenate them along the channel dimension directly. Since the computation of $Q$, $K$, and $V$ is linear, the concatenated input $[\bar{X}_{2d}|\bar{X}_{3d}]$ corresponds to $Q'$, $K'$, and $V'$ as follows:
\begin{equation}
    \left\{\begin{matrix}
        Q'=[\bar{X}_{2d}|\bar{X}_{3d}]W_q=[Q_{2d}|Q_{3d}]
        \\K'=[\bar{X}_{2d}|\bar{X}_{3d}]W_k=[K_{2d}|K_{3d}]
        \\V'=[\bar{X}_{2d}|\bar{X}_{3d}]W_v=[V_{2d}|V_{3d}]
        \\Q',K',V'\in \mathbb{R}^{N\times (d_{2d}+d_{3d})}
    \end{matrix}\right. .
\end{equation}

Our proposed MXCA can be seen as an one-step process to simultaneously obtain self-attention within individual modalities and cross-attention between multiple modalities. The specific process is shown in Eq.\eqref{mxca}.
\begin{equation}\label{mxca}
    \begin{split}
        Attention(Q',K',V')= V'\cdot Softmax({Q'}^T K'/\tau)        \\=
        V'\cdot Softmax([Q_{2d}|Q_{3d}]^T\cdot[K_{2d}|K_{3d}]/\tau) \\=
        V'\cdot Softmax\begin{pmatrix}
                           {Q_{2d}}^T K_{2d}/\tau & {Q_{2d}}^T K_{3d}/\tau
                           \\ {Q_{3d}}^T K_{2d}/\tau & {Q_{3d}}^T K_{3d}/\tau
                       \end{pmatrix}_.
    \end{split}
\end{equation} 

Similarly, regarding time complexity, this one-step process under original attention \cite{attention} is $O((N_{2d}+N_{3d})^2*d)$, while ours is $O(N*(d_{2d}+d_{3d})^2/h)$, where $N \gg d$. Therefore, our MXCA maintains lightweight properties in multimodal fusion as in 2D RGB-D fusion. 

\subsection{3D Branch}\label{sec:3d_branch}
Our 3D Branch comprises two main parts, edge convolution for feature descriptors and 3D-2D joint propagation pooling, which are illustrated in Fig.~\ref{3d_branch} (a).
\subsubsection{Edge Convolution for Feature Descriptors}
The nature of sparse depth points as 3D point clouds provides significant relative geometric relationships, which reveals a representation that differs from the local features of 2D images. Thus, the first step of our 3D branch is to aggregate this relationship between point clouds through edge convolution and obtain feature descriptors. 

Given a back-projected feature point cloud $\mathcal{P} = \{ p_i \}_{i=1}^{N_p}$, where each point $p_i$ consists of a 3D coordinate $(x_i, y_i, z_i)$, a pixel coordinate $(u_i, v_i)$, and a 24-dimensional image feature $q_{u_i, v_i}$ queried from the output of the stem module in the encoder. Thus, each point is represented as a $F$-dimensional vector, i.e., $p_i \in \mathbb{R}^F$. We apply stacked edge convolution $EdgeConv()$ layers~\cite{DGCNN} to aggregate 3D non-local neighbors of local point clouds through multiple iterations. During the early iterations, points are primarily aggregated due to close spatial distances. As iterations proceed, aggregation becomes guided more by similarity in geometric structure. 
Then, we obtain per-point descriptors $\mathcal{D} = \{d_i\}_{i=1}^{N_p}$, where $d_i \in \mathbb{R}^{64}$. For details of $EdgeConv()$, please refer to DGCNN\cite{DGCNN}.

As emphasized in NLSPN \cite{NLSPN}, non-local spatial propagation helps to solve mixed-depth problems at boundaries. Each $EdgeConv()$ layer dynamically selects k-nearest neighbors, making the 3D branch have non-local propagation properties on the point cloud modality (Fig.~\ref{3d_branch} (b-3) and (b-4)). 

\subsubsection{3D-2D Joint Propagation Pooling}
The discrete point cloud descriptors and the gridded 2D feature maps exhibit two primary gaps: the 3D-2D cross-modal representation gap, which has been addressed by forementioned MXCA in the encoder, and the sparse-dense gap, which exists between sparse point cloud features and dense image features.

To address the sparse-dense gap, we propose a 3D-2D Joint Propagation Pooling (JPP) module which enables a dense-to-dense interaction. First, to achieve 3D-2D alignment, we create an empty feature map $\mathcal{F}\in \mathbb{R}^{\frac{H}{2}\times \frac{W}{2}\times 64}$ and accumulate each descriptor $d_i$ onto the corresponding position $(\lfloor\frac{v_i}{2}\rfloor, \lfloor\frac{u_i}{2}\rfloor)$ in $\hat{\mathcal{F}}$ as Eq.~\eqref{accumulate}. Here, the operator $\lfloor \cdot  \rfloor$ represents rounding downwards.
\begin{equation}\label{accumulate}
    \hat{\mathcal{F}}(y,x,:)=\sum_{\lfloor \frac{v_i}{2} \rfloor = y,\lfloor \frac{u_i}{2} \rfloor = x} d_i,\quad 0\leqslant (y,x)<(\frac{H}{2},\frac{W}{2}).
\end{equation} 

In this process, point cloud sourced from the RGB-D image is accumulated onto the half down-sampled feature map $\hat{\mathcal{F}}$, which can be regarded as an equivalent pooling operation. 

\begin{figure*}[t]
    \captionsetup{font={small}}
    \centerline{\includegraphics[width=1.0\textwidth]{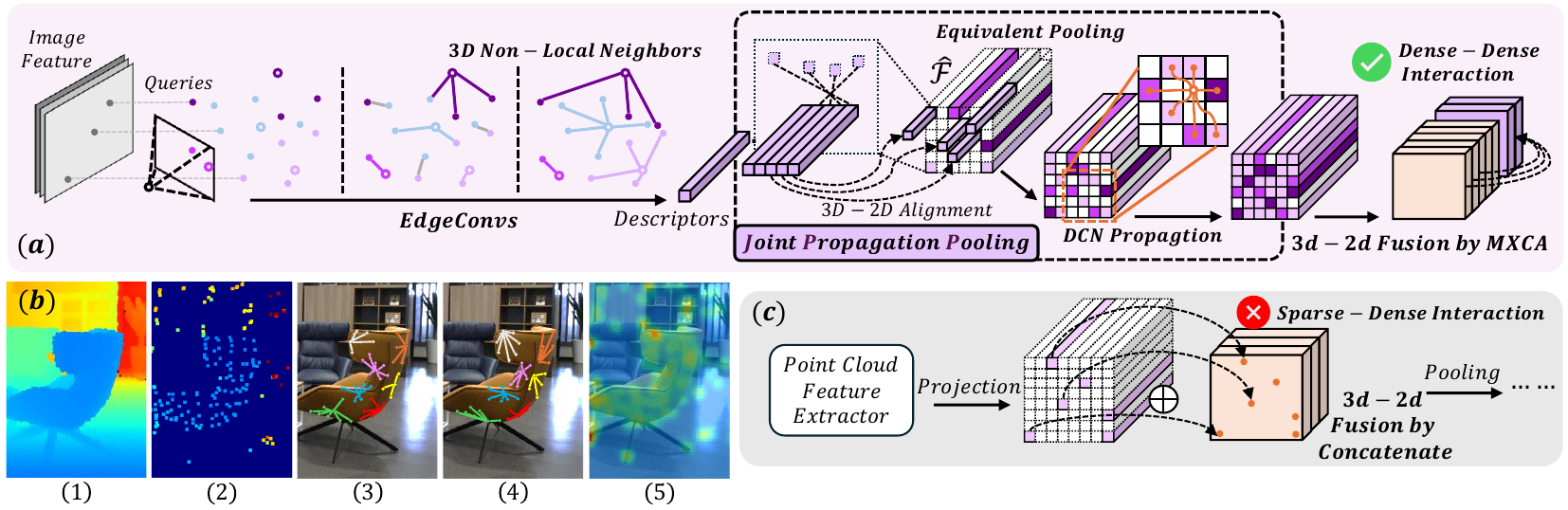}}
    \caption{\textbf{(a) Illustration of our 3D branch design details. 
    (b) Non local Propagation Properties of proposed 3D Branch:} 
    (1)-(2) Ground truth and sparse depth. (3)-(4) As the number of EdgeConv iterations 
    increases, the aggregation basis of point clouds shifts from distance to 
    geometric structure. (5) Overlayed dense activation map of $X_{3d}$.
    \textbf{(c) Previous 3D-2D fusion patterns of \cite{PointFusion, GAENet}.}}
    \label{3d_branch}
\end{figure*}

Subsequently, $\hat{\mathcal{F}}$ is normalized by a $Softmax(\cdot)$ operation along its third dimension, as there may be multiple descriptors accumulated at the same position. Then, dynamic propagation is implemented through deformable convolution \cite{DCN} (DCN), which dynamically propagate the information to the locations where point cloud descriptors are not obtained. Finally, a linear layer is used for inter-channel interaction,
thus reconstructing the feature map $X_{3d}$ of point cloud modality. The above process is formulated as follows: 
\begin{equation}
    X_{3d} = Linear(DCN(Softmax(\hat{\mathcal{F}}))) .
\end{equation}

Previous methods, such as PointFusion \cite{PointFusion} and GAENet \cite{GAENet}, project point cloud features onto the 2D plane and fuse with image features directly (Fig.~\ref{3d_branch} (c)). This projection approach results in extremely sparse interaction between the point cloud and the image pixels, with less than 1\% of the pixels contributing. By contrast, our JPP module produces dense 3D feature map for subsequent dense-to-dense interaction in MXCA.

\subsection{Decoder}\label{sec:decoder}
We designed a direct top-down decoder (Fig.~\ref{Network} bottom right) for reconstructing depth maps at three 
levels $\left\{\hat{D}_{1/1}, \hat{D}_{1/2},\hat{D}_{1/4}\right\}$. The full-resolution depth map $\hat{D}_{1/1}$ is fed into the SPN module and iterated 3 times to be refined. Our implementation of SPN module mainly follows the design of DySPN \cite{Dyspn}. One iteration of SPN could be written as 
\begin{equation}
    \begin{split}
        h_{i,j}^{t+1}=(\sum_{k\in {\mathbb{Z}_+}}\sum_{(a,b)\in
        N_{i,j,k}^{t}}\beta_1 w_{i,j}(a,b)h_{a,b}^{t}+\beta_2 h_{i,j}^{t}) \\
        \cdot(1-{C}^t)+{C}^t\beta_3 h_{i,j}^{0} , \qquad\qquad\quad
    \end{split}
\end{equation}
where $t$ represents iteration times, $h_{i,j}^{t}$ is the pixel value at $(i,j)$, and $N_{i,j,k}^{t}$ is the set of neighbors of pixel $(i,j)$ at pixel distance $k\in \mathbb{Z}_{+}$. $w_{i,j}(a,b)$ is the affinity matrix weight between pixel $(i,j)$ and its neighbour $(a,b)$. $\beta_1$,$\beta_2$ and $\beta_3$ are the weights calculated by spatial and sequential attentions. We introduced a variable confidence weight ${C}^t$, which gradually decreases with each iteration to perform confidence propagation. This is to prevent the errors of outlier depth from being retained in the final depth map. 

\subsection{Loss Function}\label{sec:loss}
We employ a combination of $\ell_1$ and $\ell_2$ loss to supervise the network training. We incorporate multi-scale weights for depth maps at different scales to assist in the early convergence of the network. Our multi-scale loss can be expressed as follows:
\begin{equation}
    L(\hat{D},D_{gt})=\sum_{s\in {\left \{\frac{1}{1}, \frac{1}{2}, \frac{1}{4} \right \}}}\frac{\gamma_s}{\left | V \right | }\sum_{(i,j)\in V}({\left | \hat{D}^s_{(i,j)}-D^{gt}_{(i,j)} \right |}^{\rho} ).
\end{equation}
Here $s$ represents different scales, $\rho=\left \{1,2 \right \}$ denotes $\ell_1$ or $\ell_2$ loss, $V$ is the index set of valid depth pixels in ground truth, and $\left | V \right |$ is the number of valid pixels. We set a weight $\gamma_s$ between scales to balance the loss during training.
\section{Dataset Evaluation}\label{sec:dataset_eval} 

\begin{figure*}[t]
    \captionsetup{font={small}}
    \centerline{\includegraphics[width=1.0\textwidth]{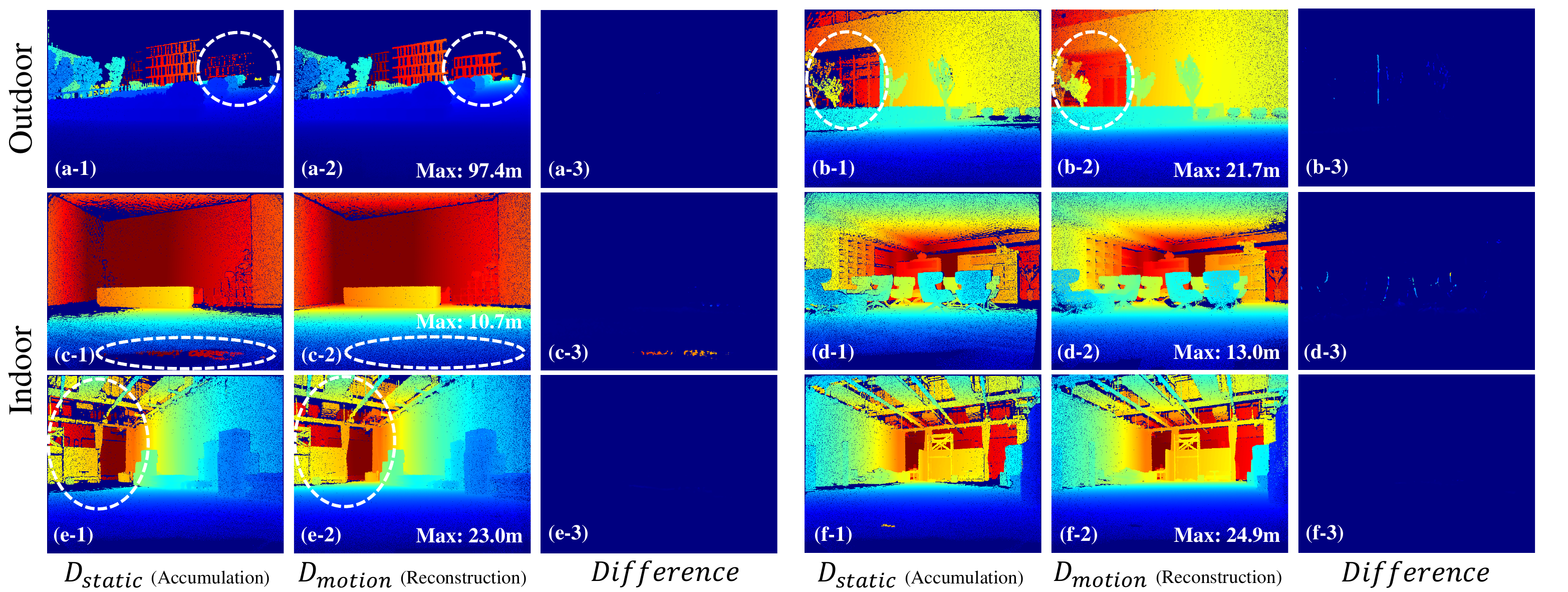}}
    \caption{\textbf{Visualized comparison of raw depth map acquired through different 
    method.} Our reconstruction-based method has significantly higher overall quality 
    compared to the accumulation-based method.} 
    \label{suppl_compare}
\end{figure*}

In this section, we validate the accuracy and return density of LASER-ToF dataset. We propose a method to directly compare depth maps acquired through accumulation-based method and our reconstruction-based method. 

\begin{figure}[th]
    \captionsetup{font={small}}
    \centerline{\includegraphics[width=0.476\textwidth]{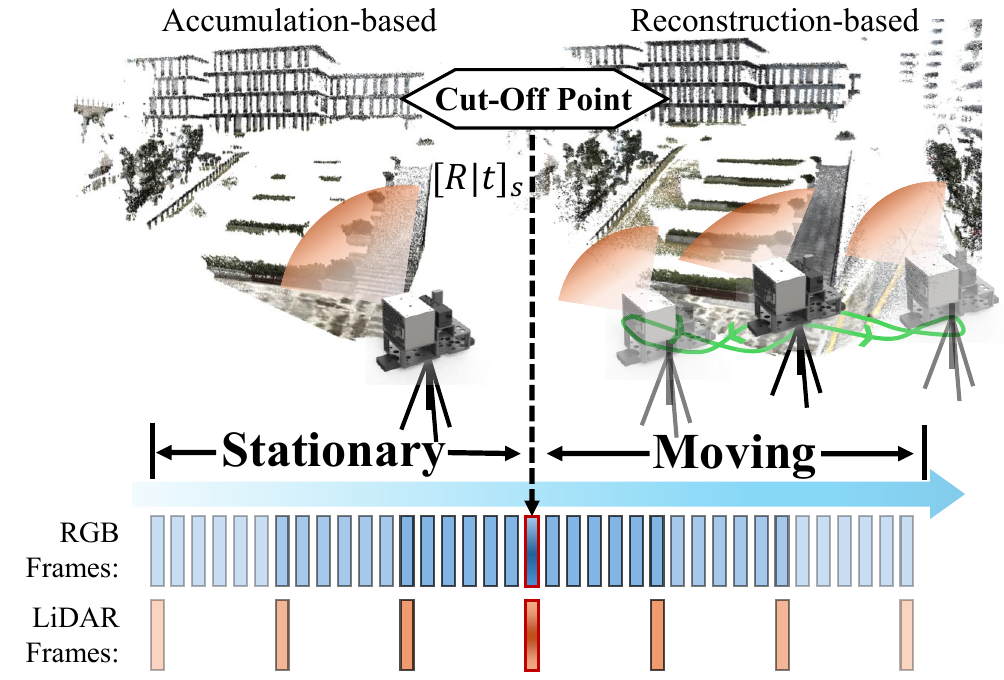}}
    \caption{\textbf{Our direct validation method,} designed specifically for comparing depth 
    map from accumulation (stationary phase) and reconstruction (moving phase). }
    \label{suppl_gt}
\end{figure}

As shown in Fig.~\ref{suppl_gt}, we first keep the handheld suite stationary for 30 seconds, then move it for 30 seconds. Taking the pose at the moment just before moving as ${[R|t]}_s$(cut-off point), we project the accumulated point cloud from the stationary phase onto the depth map at the viewpoint ${[R|t]}_s$ to obtain $D_{static}$. Similarly, we project the reconstructed point cloud from the moving phase onto the depth map at the viewpoint ${[R|t]}_s$ to obtain $D_{motion}$. In this way, we can compare the differences between the depth maps of accumulation-based and reconstruction-based methods at the cut-off point.

\begin{table}[bth]
    \centering
    \captionsetup{font={small}}
    \caption{Quantitative evaluation of $D_{motion}$ when take $D_{static}$ as ground truth and return density comparison.}
    \resizebox{0.48\textwidth}{!}{
    \begin{tabular}{l|ccc|c|c}
    \toprule
    \multirow{2}{*}{Scene} & \multicolumn{3}{c|}{$D_{static}$ as Ground Truth} & \multicolumn{2}{c}{Return Density} \\ \cline{2-6} \noalign{\vskip 0.5mm}
                           & RMSE (m) & REL      & Max Depth (m) & $D_{static}$ & $D_{motion}$      \\ 
    \midrule
    Outdoor-1              & 0.241    & 0.007    & 32.48         & 67.8\%       &  92.7\%           \\
    Outdoor-2              & 0.108    & 0.009    & 97.41         & 50.5\%       &  56.8\%           \\
    Outdoor-3              & 0.232    & 0.004    & 21.74         & 78.6\%       &  88.0\%           \\
    Outdoor-4              & 0.079    & 0.008    & 36.58         & 72.3\%       &  79.0\%           \\
    \midrule
    Indoor-1               & 0.023    & 0.002    & 24.88         & 75.4\%       &  80.7\%            \\ 
    Indoor-2               & 0.032    & 0.003    & 22.93         & 74.0\%       &  82.1\%            \\ 
    Indoor-3               & 0.046    & 0.005    & 26.97         & 72.1\%       &  73.2\%            \\ 
    Indoor-4               & 0.022    & 0.002    & 10.58         & 61.6\%       &  73.3\%            \\ 
    Indoor-5               & 0.105    & 0.006    & 23.85         & 69.0\%       &  79.1\%            \\ 
    Indoor-6               & 0.240    & 0.006    & 12.99         & 74.8\%       &  82.6\%            \\ 
    Indoor-7               & 0.103    & 0.008    & 11.19         & 70.4\%       &  79.9\%            \\ 
    \midrule
    Average                & 0.112    & 0.006    & 27.69         & 70.0\%       &  79.4\%            \\ 
    \bottomrule
    \end{tabular}
    }

    \label{compare_gt}

\end{table}

Under this setting, we evaluate our reconstruction-based dataset collection method in eleven outdoor and indoor scenes. Fig.~\ref{suppl_compare} visualizes the comparison between $D_{static}$ and $D_{motion}$ in several selected scenes. We find that, compared to $D_{static}$, $D_{motion}$ can effectively avoid missing regions, provide denser depth, and maintain the same fine-grained structure as $D_{static}$. In contrast, $D_{static}$ is susceptible to reflective surfaces, \textit{i.e. multipath effect} (Fig.~\ref{suppl_compare}, (c-1)), resulting in incorrect depth regions. 

In Table~\ref{compare_gt}, we evaluate the quantitative accuracy of $D_{motion}$ using 
$D_{static}$ as reference ground truth. The metrics evaluated in Table~\ref{compare_gt} include root mean square error (RMSE), relative mean absolute error (REL) and return density, which are defined as follows:
\begin{equation*}
\operatorname{RMSE}(\hat{D}, D^{\text{gt}}) = \sqrt{ \frac{1}{HW} \cdot \sum_{i,j}^{W,H} (\hat{D}_{i,j} - D^{\text{gt}}_{i,j})^2} ,
\end{equation*}
\begin{equation*}
\operatorname{REL}(\hat{D}, D^{\text{gt}}) = \frac{1}{HW} \cdot \sum_{i,j}^{W,H} \frac{|\hat{D}_{i,j} - D^{\text{gt}}_{i,j}|}{D^{\text{gt}}_{i,j}} ,
\end{equation*}
\begin{equation*}
\operatorname{Return\ Density}(D)=\frac{1}{HW}\sum_{i,j}^{W,H}\mathcal{I}\{D_{i,j} > 0\},
\end{equation*}
where $\hat{D}$ denotes the depth map to be evaluated, $D^{\text{gt}}$ denotes the reference depth map, $\mathcal{I}(\cdot)$ denotes indicator function, and $H$ and $W$ are the height and width of depth maps, respectively.

For eleven scenes with an average maximum depth of 27.69\,m, the average REL of $D_{motion}$ is only 0.6\%, indicating slight differences between $D_{motion}$ and $D_{static}$ in the valid regions of $D_{static}$. In addition, the return density of reconstruction-based $D_{motion}$ is 9.4\% higher than that of $D_{static}$.

In conclusion, Fig.~\ref{suppl_compare} and Table~\ref{compare_gt} demonstrates that $D_{motion}$ is a reasonable representation of ground truth with higher return density. 
Here we just compare the raw depth map without any post processing. In practice, 
we apply a minimum filter (Section~\ref{subsec:gt}) to $D_{motion}$ to remove 
invisible background points and further improve return density to 94.6\%. 
Moreover, reconstruction-based methods are significantly more efficient than accumulation-based methods, as the former can leverage all historical point clouds to consecutively project and obtain depth maps from various viewpoints. During the dataset collection process, 1-2 minutes of scanning can produce data from 300-500 viewpoints in a scene.

\begin{table*}[th]
    \centering
    \renewcommand\arraystretch{1.02}
    \captionsetup{font={small}}
    \caption{Quantitative results on LASER-ToF benchmark. (\colorbox{mygreen!75}{\phantom{xx}}: best; \colorbox{mygreen!35}{\phantom{xx}}: second-best. Horizontal group 1 and 3: supervised methods. Horizontal group 2: generalizable methods.)}
    \resizebox{0.96\textwidth}{!}{
        \begin{tabular}{l|rrcc|rrcc|r|r}
            \toprule
                                            & \multicolumn{4}{c|}{RGB with ToF-Only as Input} & \multicolumn{4}{c|}{RGB with ToF\&Visual as Input}  & &  \\ \cline{2-9} \noalign{\vskip 0.5mm}
            Method                           & \makecell[c]{RMSE$\downarrow$\\(mm)} & \makecell[c]{MAE$\downarrow$\\(mm)}  & REL$\downarrow$ & $\delta_1\uparrow $  & \makecell[c]{RMSE$\downarrow$\\(mm)} & \makecell[c]{MAE$\downarrow$\\(mm)} & REL$\downarrow$ & $\delta_1\uparrow $ & \makecell[c]{Params.$\downarrow$\\(M)} & \makecell[c]{FLOPs$\downarrow$\\(G)} \\
            \midrule
            NLSPN \cite{NLSPN}              & 1140.49 & \cellcolor{mygreen!35}496.24 & \cellcolor{mygreen!35}0.0605   & \cellcolor{mygreen!35}94.7  & 1033.02          & \cellcolor{mygreen!35}412.91           & \cellcolor{mygreen!35}0.0530          & \cellcolor{mygreen!35}95.7                 & 26.23             & 972.48    \\
            MDANet \cite{MDANet}            & 1212.54 & 619.98 & 0.0791  & 93.9   & 1102.83          & 521.72                                 & 0.0725          & 94.2                 & \cellcolor{mygreen!75}3.04              & \cellcolor{mygreen!75}323.52 \\
            PENet \cite{PENet}              & 1105.80 & 528.64 & 0.0674  & 94.5   & \cellcolor{mygreen!35}949.98           & 421.33           & 0.0582          & \cellcolor{mygreen!35}95.7                 & 131.92            & 585.61    \\
            DySPN \cite{Dyspn}              & 1170.58 & 547.37 & 0.0701  & 94.1   & 1058.74          & 467.00                                 & 0.0634          & 94.8                  & 26.80             & 934.18    \\
            CFormer \cite{CompletionFormer} & \cellcolor{mygreen!35}1086.06 & 526.65 & 0.0694  &  94.2   & 987.86           & 447.64                                  & 0.0630          & 95.0               & 82.51             & 769.88    \\
            LRRU \cite{LRRU}                & 1531.94 & 845.39 & 0.1101  & 89.0    &   1200.87          & 616.30                                       & 0.0849          & 93.3                      & 20.84             & 1294.98    \\
            DFU \cite{DFU}                  & 1570.33 & 864.10 & 0.1147  & 87.6    &   1205.85          & 628.50                                        & 0.0954          & 90.9                    & 25.47              & 1164.62   \\ 
            BP-Net \cite{BP-Net}            & 1490.85 & 778.22 & 0.1056  & 90.3    &   1180.48          & 568.01                                       & 0.0784          & 93.0                  & 89.87             & 1032.98   \\ \noalign{\vskip 0.7mm}
            \textbf{Average}                & 1288.57 & 650.82 & 0.0846  & 92.3    &   1089.95          & 510.43                                        & 0.0711          & 94.1                    & 54.02              & 884.78   \\ 
            \midrule
            OMNI-DC \cite{zuo2025omni}      & 3904.18 & 2002.56 & 0.1739 & 70.1   & 1902.78          & 889.81                               & 0.1136          & 88.5                   & 416.84            & 2349.06   \\
            Marigold-DC~\cite{viola2025marigold} & 3950.21 & 2034.30 & 0.1802 & 69.8   & 2719.61          & 1321.26                             & 0.1775          & 82.4                       &  $>$1000            & $>$3000  \\
            \midrule
            \textbf{ToFormer (ours)}        & \cellcolor{mygreen!75}1024.08 & \cellcolor{mygreen!75}453.69 & \cellcolor{mygreen!75}0.0575  & \cellcolor{mygreen!75}95.4  & \cellcolor{mygreen!75}924.07  & \cellcolor{mygreen!75}379.06   & \cellcolor{mygreen!75}0.0501 & \cellcolor{mygreen!75}96.2          & \cellcolor{mygreen!35}7.61              & \cellcolor{mygreen!35}507.49    \\
            \bottomrule
        \end{tabular}
    }

    \label{metrics_laser}

\end{table*}

\section{Benchmark and Ablation Study}
\subsection{Benchmark Setup}
\textbf{LASER-ToF Depth Benchmark.} LASER-ToF is collected by our multi-sensor platform in section \ref{sec:dataset}, which intends for challenging large-scale indoor/outdoor ToF depth completion. It contains 20,996 sets, each set of data is composed of RGB image, sparse depth (ToF-only), sparse depth (ToF\&Visual), and ground truth depth with a resolution of 640$\times$480. There are two options for sparse depth: ToF-Only or ToF\&Visual Depth (ToF with additional sparse depth from visual SLAM), to evaluate model performance under different sparse depth conditions. We split the dataset into 18,746 training images, 750 validation images, and 1,500 test images (for benchmark). The test set consists of samples from various unseen scenes and perspectives. 

\textit{Benchmark methods:} We follow the original implementation for each supervised methods, including NLSPN \cite{NLSPN}, MDANet \cite{MDANet}, PENet \cite{PENet}, DySPN \cite{Dyspn}, CFormer (CompletionFormer) \cite{CompletionFormer}, LRRU \cite{LRRU}, DFU \cite{DFU}, and BP-Net \cite{BP-Net}. Each method is first validated under its original scenario (simulated uniform depth sampling) to ensure faithful reproduction. Then, all supervised methods and ToFormer are trained from scratch on LASER-ToF training set with data augmentation including gaussian noise and random masks on sparse depth map. The training set mixes ToF-only, ToF\&Visual, and augmented sparse depth maps. 
After training, the models are evaluated under two sparse depth patterns: ToF-Only and ToF\&Visual, respectively.

In addition to supervised baselines, we also evaluate the recent generalizable depth completion models OMNI-DC~\cite{zuo2025omni} (416.84M parameters, trained on 573K samples) and Marigold-DC~\cite{viola2025marigold} ($>$1000M parameters, pretrained and fine-tuned on more than 2.3 billion samples), both of which have large parameter counts and are trained across multiple datasets. Fine-tuning such models on our domain-shifted train set would be computationally expensive and may also degrade its pretrained generalization ability. Therefore, following standard practice, we evaluate OMNI-DC and Marigold-DC in a zero-shot manner using the official pretrained weights, as they are designed for cross-domain generalization. Their results are reported separately from fully supervised methods to avoid misleading comparisons.

\textbf{NYU-Depth v2 Benchmark~\cite{NYUv2}.} 
Besides large-scale scenario ToF depth sampling, we also evaluate our method under the setting of random uniform depth sampling. We follow the same setting of previous methods~\cite{Sparse-to-dense,NLSPN,CompletionFormer}, training our model on 50k images sampled from the training set and test on the 654 images from the offical processed test set. Original frames of resolution 640$\times$480 are half down-sampled and then center-cropped to 304$\times$228. 500 depth points are randomly sampled from the dense ground truth as sparse depth. We directly report the numbers provided in the original paper for all competitor methods. 

\subsection{Implementation Details}
We implement our model in PyTorch framework \cite{pytorch} on a NVIDIA L20 GPU. We adopt AdamW optimizer with an initial learning rate of 0.00015, $\beta_1$=0.9, $\beta_2$=0.999 and weight decay of 0.01. On both LASER-ToF and NYUv2 datasets, we train the model for 80 epochs and decay the learning rate by a factor of 0.5 at epochs 28, 36, 48, 60, 72. 

\begin{figure*}[t]
    \captionsetup{font={small}}
    \centerline{\includegraphics[width=1.0\textwidth]{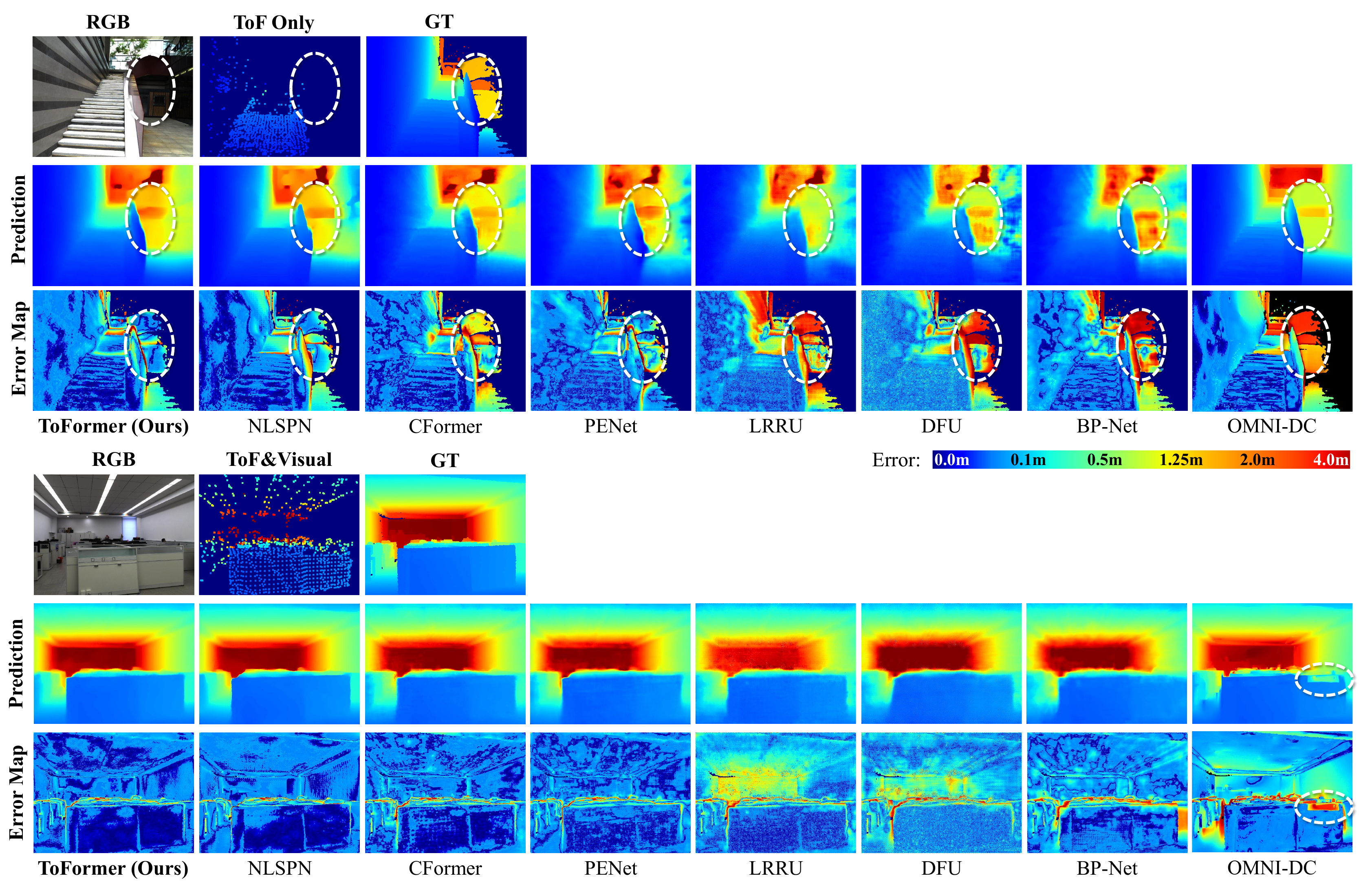}}

    \caption{\textbf{Qualitative results on the LASER-ToF benchmark.} \textbf{Upper part:} An outdoor staircase scene, where only ToF is used as sparse depth input. Depth is available only for the few steps directly in front of the sensor. \textbf{Lower part:} An office scene, near-range and far-range depths are provided by ToF and visual SLAM respectively, yet the textureless white wall yields no valid depth points. }
    \label{visual_0}

\end{figure*}

For quantitative evaluation, we follow the standard metrics \cite{NYUv2,Sparse-to-dense,cspn}: root mean squared error (RMSE [mm]), mean absolute error (MAE [mm]), relative mean absolute error (REL), and percentages $\delta_n$ of inlier pixels. The metrics are defined as follows:
\begin{equation*}
\operatorname{RMSE}(\hat{D}, D^{\text{gt}}) = \sqrt{ \frac{1}{HW} \cdot \sum_{i,j}^{W,H} (\hat{D}_{i,j} - D^{\text{gt}}_{i,j})^2} ,
\end{equation*}
\begin{equation*}
\operatorname{MAE}(\hat{D}, D^{\text{gt}}) = \frac{1}{HW} \cdot \sum_{i,j}^{W,H} |\hat{D}_{i,j} - D^{\text{gt}}_{i,j}| ,
\end{equation*}
\begin{equation*}
\operatorname{REL}(\hat{D}, D^{\text{gt}}) = \frac{1}{HW} \cdot \sum_{i,j}^{W,H} \frac{|\hat{D}_{i,j} - D^{\text{gt}}_{i,j}|}{D^{\text{gt}}_{i,j}}, 
\end{equation*}
\begin{equation*}
\delta_n(\hat{D}, D^{\text{gt}}) = \frac{1}{HW} \sum_{i,j}^{W,H} \mathcal{I} \{ \max \left( \frac{\hat{D}_{i,j}}{D^{\text{gt}}_{i,j}}, \frac{D^{\text{gt}}_{i,j}}{\hat{D}_{i,j}} \right) < 1.25^n \} ,
\end{equation*}
where $\hat{D}$ denotes predicted depth, $D^{\text{gt}}$ denotes ground-truth depth, $\mathcal{I}(\cdot)$ denotes indicator function, and $H$ and $W$ are the height and width of depth maps, respectively.

\begin{figure*}[ht]
    \captionsetup{font={small}}
    \centerline{\includegraphics[width=1.0\textwidth]{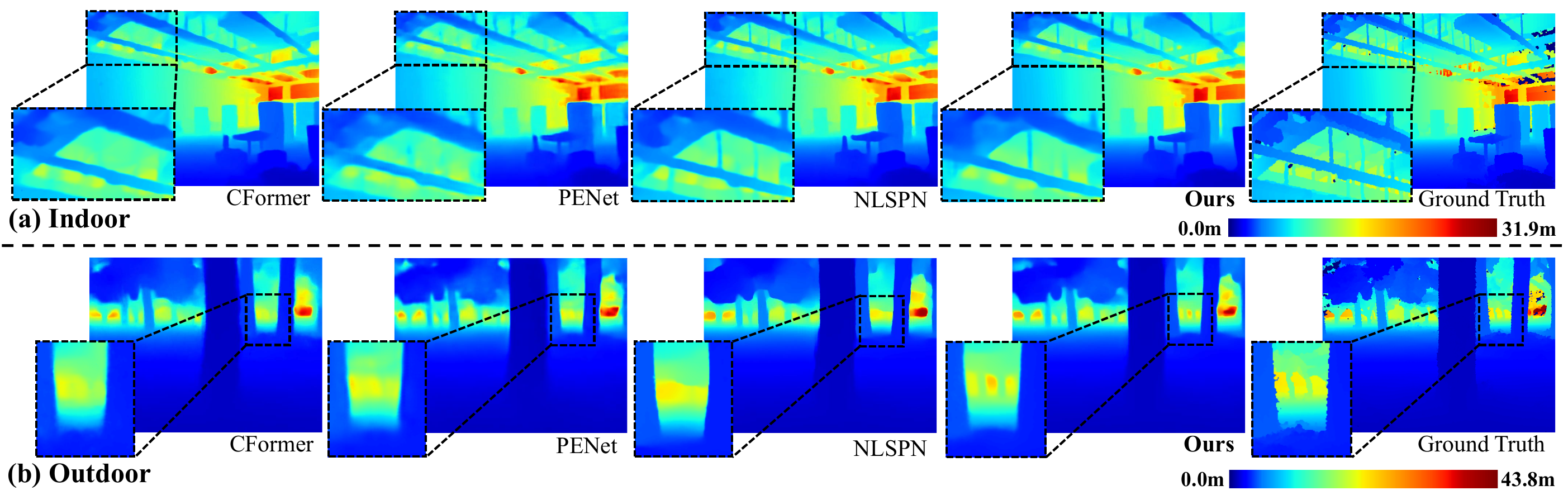}}

    \caption{\textbf{Visualized comparison of geometric details for four best methods in indoor and outdoor scenarios.} The zoomed-in regions show fine-grained differences. }
    \label{visual_1}

\end{figure*}

\begin{figure}[th]
    \captionsetup{font={small}}
    \centerline{\includegraphics[width=0.48\textwidth]{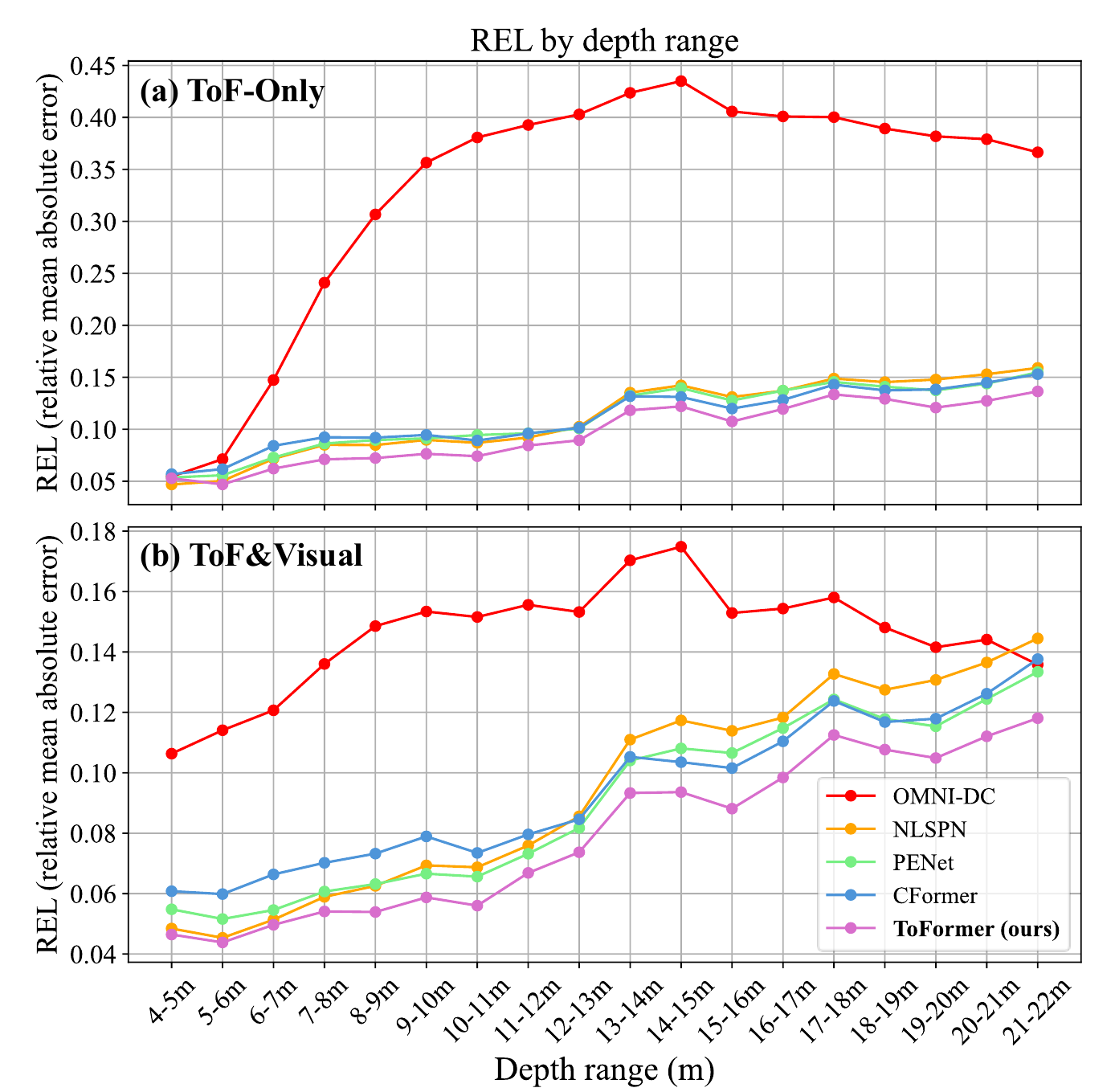}}
    \vspace{-5pt}
    \caption{\textbf{Relative mean absolute error (REL) across depth ranges.} As the depth increases toward far range, all methods exhibit a rising error trend. Our method maintains the lowest error across nearly all depth ranges. }
    \label{rel_range}
\end{figure}

\subsection{Comparison with Previous Methods}\label{sec:comparison} 

\begin{figure*}[ht]
    \captionsetup{font={small}}
    \centerline{\includegraphics[width=1.0\textwidth]{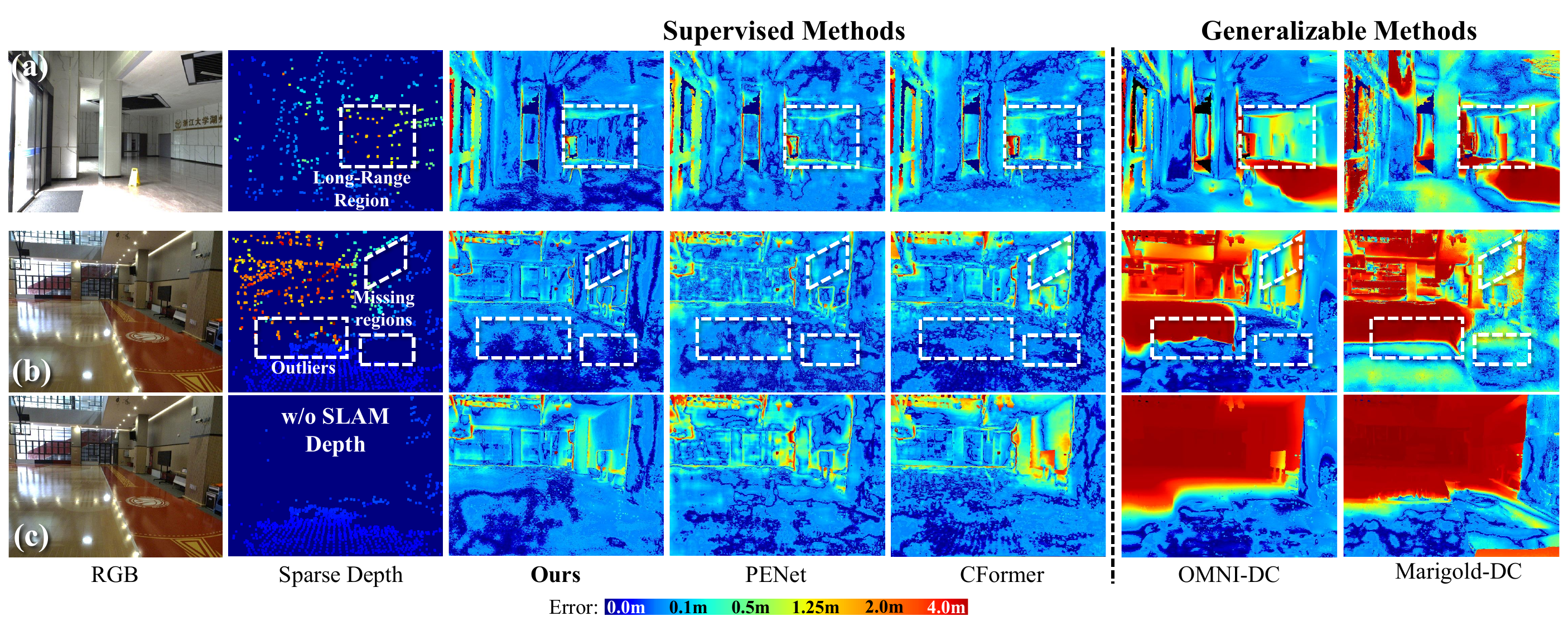}}

    \caption{\textbf{Error maps comparison.} (a) A large hall scene in which the far-range depth is provided by visual SLAM but contains noise. (b) A challenging scene involving outliers caused by ground reflections and multiple large missing regions. (c) The same scene as in (b) when only ToF depth is used and SLAM depth is unavailable.}
    \label{Error}

\end{figure*}

\textbf{Quantitative results.} Table~\ref{metrics_laser} shows the quantitative results of LASER-ToF benchmark, where the first horizontal group and our ToFormer are supervised methods, the second horizontal group are generalizable methods. Dark green and light green represent the best and second-best, respectively. Our method demonstrates a significant lead across various metrics while maintaining relatively low computational overhead (second only to MDANet~\cite{MDANet}). 

Compared to the second-best methods, when using ToF-Only as sparse depth, our ToFormer shows 5.7\%, 8.6\%, and 5.0\% reductions of RMSE, MAE and REL. While for ToF\&Visual cases, our ToFormer shows 2.7\%, 8.2\%, and 5.5\% reductions of RMSE, MAE and REL. In addition, we report the average performance of all competing supervised methods as a reference to better reflect the overall performance level. Compared to this average baseline, our ToFormer achieves substantially larger improvements. Specifically, under the ToF-Only setting, our method reduces RMSE, MAE, and REL by 25.8\%, 43.5\%, and 47.1\%, respectively. Under the ToF\&Visual setting, the corresponding reductions are 17.6\%, 34.7\%, and 41.9\%. 

We observe that multi-level iterative methods including LRRU~\cite{LRRU}, DFU~\cite{DFU}, and BP-Net~\cite{BP-Net}, do not exhibit a performance advantage on the LASER-ToF benchmark. This may be related to their reliance on uniform and accurate sparse depth.  The RMSE metrics of OMNI-DC~\cite{zuo2025omni} and Marigold-DC are more than twice those of supervised methods. This reveals the limitation of existing generalizable methods to handle ToF depth completion in large-scale scenarios.

Fig.~\ref{rel_range} compares the relative mean absolute error (REL) of the four best supervised methods (our proposed ToFormer, CFormer~\cite{CompletionFormer}, PENet~\cite{PENet}, NLSPN~\cite{NLSPN}) and one generalizable method (OMNI-DC~\cite{zuo2025omni}) across different depth ranges. As the distance increases, all methods show an upward trend. Our ToFormer maintains the lowest REL in almost every range whether using ToF-Only or ToF\&Visual as sparse depth. While NLSPN~\cite{NLSPN} exhibits comparable performance to our ToFormer within the 4–7\,m range, it gradually fall behind ToFormer as distance increases. The generalizable method OMNI-DC~\cite{zuo2025omni} exhibits substantially higher REL than our ToFormer. The gap between OMNI-DC and ToFormer widens within the 4–15\,m range, but the gap is slightly smaller when the range extends to farther regions (e.g., beyond 17m).

\textbf{Qualitative results.} Qualitative results on the LASER-ToF benchmark are provided in Fig.~\ref{visual_0}, Fig.~\ref{visual_1}, and Fig.~\ref{Error}. Fig.~\ref{visual_0} presents the predicted depth maps and error maps of various methods under  ToF-only and ToF\&Visual settings. In the error maps, warmer colors indicate larger errors. From these results, our ToFormer demonstrates strong capability in accurately completing regions with available sparse depth, while also leveraging spatial geometric relationships to infer dense depth in areas without any depth observations. LRRU \cite{LRRU}, DFU \cite{DFU}, and BP-Net \cite{BP-Net} struggle when handling large missing regions. This phenomenon aligns with the limitations of ``dealing long-range information delivery'' discussed in the original BP-Net \cite{BP-Net} paper. Although OMNI-DC~\cite{zuo2025omni} learns strong priors from large-scale data and preserves sharp edge details, it exhibits substantial absolute errors when sparse depth is missing.

Fig.~\ref{visual_1} compares the geometric details of the four methods with the highest quantitative accuracy in indoor and outdoor scenarios. Our ToFormer and NLSPN~\cite{NLSPN} exhibit similar superiority in distinguishing fine structures between the foreground and background (Fig.~\ref{visual_1} (a)). This may be attributed to the non-local propagation design in these models, which helps to solve mixed-depth problems at boundaries. In addition, in the outdoor forest scene of Fig.~\ref{visual_1} (b), our ToFormer can effectively maintain geometric details in the distance.  

Fig.~\ref{Error} compares the error maps of three best supervised methods and two generalizable methods under two scenarios and different sparse depth setup. In Fig.~\ref{Error} (a), the sparse depth provided by visual SLAM in long-range region is noisy. Our ToFormer preserves better accuracy under noisy long-range depth inputs. Fig.~\ref{Error} (b) illustrates a challenging depth completion case involving outliers caused by ground reflections and large missing regions. Our method demonstrates superior robustness over PENet~\cite{PENet} and CFormer~\cite{CompletionFormer}, whereas the generalizable methods OMNI-DC~\cite{zuo2025omni} and Marigold-DC~\cite{viola2025marigold} are heavily misled by reflective surfaces and outliers, producing large erroneous regions. Fig.~\ref{Error} (c) shows ToF-Only situation without SLAM depth. Although our method exhibits a slight increase in error, it remains the best-performing approach. In contrast, the generalizable methods degrade severely under this condition and fail to effectively complete depth in large missing regions.

\textbf{Results under uniform sampling.} We also report a quantitative comparison on the NYUv2 dataset \cite{NYUv2} in Table~\ref{metrics_nyu}. This experiment evaluates the performance of our method under the previous uniform-sampled sparse depth. The upper half presents the results of 2D-based methods, while the lower half shows the results of 2D-3D joint methods. Although not specifically designed for uniform depth sampling, our method still achieves competitive performance under this setting, suggesting that it does not overfit to the sensor-specific assumption of ToF cameras.

\begin{table}[h]
    \centering
    \renewcommand\arraystretch{1.02}
    \captionsetup{font={small}}
    \caption{Quantitative evaluation under uniform sparse depth setting on NYUv2. }
    \resizebox{0.44\textwidth}{!}{
        \begin{tabular}{l|cc|ccc}
            \toprule
            {Method}                       & \makecell[c]{RMSE$\downarrow$\\(m)}& REL$\downarrow$ & $\delta_1\uparrow $ & $\delta_2\uparrow $ & $\delta_3\uparrow $ \\
            \midrule
            CSPN \cite{cspn}                & 0.117            & 0.016           & 99.2                & 99.9                & 100.0               \\
            GuideNet \cite{GuideNet}        & 0.101            & 0.015           & 99.5                & 99.9                & 100.0               \\
            NLSPN \cite{NLSPN}              & 0.092            & 0.012           & 99.6                & 99.9                & 100.0               \\
            CFormer \cite{CompletionFormer} & 0.091            & 0.012           & 99.6                & 99.9                & 100.0               \\
            DySPN \cite{Dyspn}              & 0.090            & 0.012           & 99.6                & 99.9                & 100.0               \\
            BP-Net \cite{BP-Net}            & 0.089            & 0.012           & 99.6                & 99.9                & 100.0               \\
            OMNI-DC \cite{zuo2025omni}            & 0.111            & 0.014           & 99.4             & 99.9             & 100.0          \\
            \midrule
            GAENet \cite{GAENet}            & 0.114            & 0.018           & 99.3                & 99.9                & 100.0               \\
            ACMNet \cite{ACMNet}            & 0.105            & 0.015           & 99.4                & 99.9                & 100.0               \\
            PRNet \cite{PRNet}              & 0.104            & 0.014           & 99.4                & 99.9                & 100.0               \\
            DFU \cite{DFU}                  & 0.091            & 0.011           & 99.6                & 99.9                & 100.0               \\
            PointFusion \cite{PointFusion}  & 0.090            & 0.014           & 99.6                & 99.9                & 100.0               \\
            \textbf{ToFormer (ours)}        & 0.095            & 0.013           & 99.5                & 99.9                & 100.0               \\
            \bottomrule
        \end{tabular}
    }

    \label{metrics_nyu}

\end{table}

\subsection{Ablation Studies and Analysis}

\subsubsection{Effect of Proposed Modules} To assess the impact of the main components of ToFormer, we choose our basic encoder-decoder model with 2D RGB-D fusion as baseline, and conduct ablation experiments on LASER-ToF dataset. Results are reported in Table~\ref{ablation_self} and Table~\ref{ablation_3d}. 

\textbf{3D Branch and Joint Propagation Pooling (JPP).} Our 3D branch aggregates non-local neighbors in the point cloud structure and establish dense-to-dense interactions with 2D features via the JPP module. Compared to models without the 3D branch (B and E), the inclusion of the 3D branch (D and F) reduces the REL by 3.0\% and 4.0\%, respectively. In Table~\ref{ablation_3d}, we further investigate the effects of the JPP module. Beyond our expectation, JPP module affects performance significantly. JPP module provides dense-to-dense interaction and reduces RMSE by 50.25 mm compared to conventional sparse-to-dense interaction. 

\begin{table}[th]
    \centering
    \captionsetup{font={small}}
    \caption{Ablation studies of ToFormer's main components on LASER-ToF Depth Dataset.}
    \resizebox{0.476\textwidth}{!}{
        \begin{tabular}{l|cccc|cc|cc}
            \toprule
                & Base         &  SPN         & MXCA         & 3D              & \makecell[c]{RMSE$\downarrow$\\(mm)}& REL$\downarrow$ & GFLOPs$\downarrow$ \\
            \midrule
            (A) & $\checkmark$ & $\times$     & $\times$     & $\times$        & 957.15           & 0.0541          & 417.73        \\
            (B) & $\checkmark$ & $\checkmark$ & $\times$     & $\times$        & 950.87           & 0.0526          & 429.82        \\
            (C) & $\checkmark$ & $\times$     & $\checkmark$ & $\checkmark$    & 942.23           & 0.0525          & 495.40        \\
            (D) & $\checkmark$ & $\checkmark$ & $\times$     & $\checkmark$    & 935.15           & 0.0510          & 483.90        \\
            (E) & $\checkmark$ & $\checkmark$ & $\checkmark$ & $\times$        & 936.39           & 0.0522          & 440.52        \\
            \midrule
            (F) & $\checkmark$ & $\checkmark$ & $\checkmark$ & $\checkmark$    & 924.07           & 0.0501          & 507.49        \\
            \bottomrule
        \end{tabular}
    }
    \label{ablation_self}

\end{table}

\begin{table}[th]
    \centering
    \captionsetup{font={small}}
    \caption{Ablation studies of the 3D branch without and with JPP module. }
    \resizebox{0.476\textwidth}{!}{
    \begin{tabular}{l|c|c}
    \toprule
              & \multicolumn{2}{c}{3D Branch} \\
    \midrule
    Interaction   & Sparse-to-Dense (w/o JPP) & Dense-to-Dense (w/ JPP) \\ 
    \midrule
    RMSE (mm) & 974.32 & \textbf{924.07}  \\
    \bottomrule
    \end{tabular}
    }
    \label{ablation_3d}

\end{table}

\textbf{Multimodal Cross-Covariance Attention (MXCA).} The MXCA module is located in stage 1 of the encoder and enables early fusion of three modalities:
RGB, Sparse Depth, and Point Cloud. In \cref{ablation_self}, compared to the version without the MXCA module, the introduction of MXCA brings a slight additional computational cost but significantly improves performance metrics. For instance, when Model (B) incorporates the MXCA to become Model (E), the RMSE decrease by 14.48 mm, while the FLOPs only increase by 10.7G. 

\textbf{SPN Module} is adopted to refine the local details of final depth map. 
When MXCA and 3D branch are not included in the network architecture, the SPN module can reduce RMSE by 6.28mm from (A) to (B). When the SPN module is operated together with MXCA and 3D branch, the improvement in accuracy is more significant, with a decrease of 18.16mm in RMSE from (C) to (F). This indicates that our proposed MXCA and 3D branch provide a reliable affinity matrix and initial depth prediction for the final SPN module. 

\subsubsection{Discussion on SLAM Systems}
Our model exhibits better performance when using ToF\&Visual as sparse depth input, which may be influenced by visual SLAM. Here, we conduct ablation studies to reveal the effects of visual SLAM system to our completion network in various cases. We perform zero-shot testing on the same checkpoint in Table~\ref{metrics_laser} to evaluate the performance under multiple SLAM's types or behaviors. 

\textbf{Different SLAM and keypoint styles.} The upper half of Table~\ref{ablation_slam}
demonstrate that our model exhibits robustness to different SLAM and keypoint types. 
Without any fine-tuning, our model can directly adapt to different SLAM systems 
while maintaining competitive accuracy. 

\textbf{What if SLAM fails or degrades?} 
During SLAM operation, the system may encounter degenerate scenarios, such as textureless regions or illumination changes, which directly lead to a reduction in the number of trackable map points or possible tracking failure. 
We evaluated various potential SLAM behaviors, including tracking degradation, absence of backend, and complete SLAM failure. As shown in the lower half of Table~\ref{ablation_slam}, we found that backend optimization has the most significant impact on accuracy, as its absence diminishes both the quantity and quality of the visual point cloud. While tracking degradation impacts less on accuracy. When SLAM fails, the model can still leverage the reliable short-range depth of ToF camera to perform depth completion, though the cost is a reasonable decrease in accuracy.

\begin{table}[h]
    \centering
    \captionsetup{font={small}}
    \caption{Ablation studies on the types and behaviors of SLAM.}
    \resizebox{0.47\textwidth}{!}{
        \begin{tabular}{l|l|cc}
            \toprule
            SLAM method                          &  Keypoint                                 &  \makecell[c]{RMSE$\downarrow$\\(mm)} & \makecell[c]{MAE$\downarrow$\\(mm)}      \\
            \midrule
            ORB-SLAM \cite{ORB-SLAM3}           & Oriented FAST \cite{orb_feature}                  & 924.07                            & 379.06     \\
            \midrule
            Vins-Mono \cite{Vins}               & Shi-Tomasi \cite{gftt}                            & 1025.37                           & 435.45     \\
            \midrule
            Superpoint-SLAM \cite{sp-slam}     & Superpoint \cite{superpoint}                       & 1000.62                           & 408.71     \\
            \midrule
            \multicolumn{2}{c|}{No backend optimization}                                             & 1068.61                          & 540.42     \\
            \multicolumn{2}{c|}{Tracking degradation (50\% visual points)}                              & 946.88                           & 393.34    \\
            \multicolumn{2}{c|}{Tracking degradation (20\% visual points)}                              & 974.10                           & 422.63   \\
            \multicolumn{2}{c|}{No SLAM input (0\% visual points)}                                   & 1024.08                          & 453.69   \\
            \bottomrule
        \end{tabular}
    }
    
    \label{ablation_slam}

\end{table}

\begin{table}[h]
    \centering
    \captionsetup{font={small}}
    \caption{Computational cost comparison with previous methods at 320$\times$240 resolution.}
    \label{cost}
    \begin{threeparttable}
    \resizebox{0.48\textwidth}{!}{
        \begin{tabular}{l|rrr|r}
            \toprule
            {Method}                       
            & \makecell[c]{Params.$\downarrow$\\(M)} 
            & \makecell[c]{FLOPs$\downarrow$\\(G)}  
            & \makecell[c]{Runtime$\downarrow$\\(ms)} 
            & \makecell[c]{RMSE$\downarrow$\\(mm)}  \\
            \midrule
            NLSPN\cite{NLSPN}              
            & 26.23                              & 219.01                        & \cellcolor{mygreen!75}22.6  & 1033.02          \\
            MDANet\cite{MDANet}            
            & \cellcolor{mygreen!75}3.04         & \cellcolor{mygreen!75}86.27   & 134.7                     & 1102.83        \\
            PENet\cite{PENet}              
            & 131.92                             & 156.16                        & 329.8                     & \cellcolor{mygreen!35}949.98     \\
            DySPN\cite{Dyspn}              
            & 26.80                              & 233.55                        & 55.4                      & 1058.74      \\
            CFormer\cite{CompletionFormer} 
            & 82.51                              & 172.08                        & 159.6                     & 987.86     \\
            LRRU\cite{LRRU}                
            & 20.84                              & 323.74                        & 45.5                      & 1200.87     \\
            DFU\cite{DFU}                  
            & 25.47                              & 291.15                       & 120.7                     & 1205.85      \\
            BP-Net\cite{BP-Net}            
            & 89.87                              & 275.46                        & 218.0                     & 1180.48    \\
            OMNI-DC\cite{zuo2025omni}      
            & 416.84                             & 1750.58                       & 195.5                     & 1902.78    \\ 
            \noalign{\vskip 0.7mm}
            Average\tnote{*}               
            & 54.02                              & 219.68                       & 135.8                     & 1089.95    \\
            \midrule
            ToFormer (RTX3090)              
            & \cellcolor{mygreen!35}7.61         & \cellcolor{mygreen!35}112.84  & \cellcolor{mygreen!35}35.6 & \cellcolor{mygreen!75}924.07     \\
            ToFormer (Orin NX)             
            & 7.61                               & 112.84                        & 106.2                     & 924.07    \\ 
            \bottomrule
        \end{tabular}
    }
    \begin{tablenotes}
        \footnotesize
        \item[*] OMNI-DC is excluded from the average statistics due to its substantially larger model size and computational cost.
    \end{tablenotes}
    \end{threeparttable}
\end{table}

\begin{figure*}[th]
    \captionsetup{font={small}}
    \centerline{\includegraphics[width=1.0\textwidth]{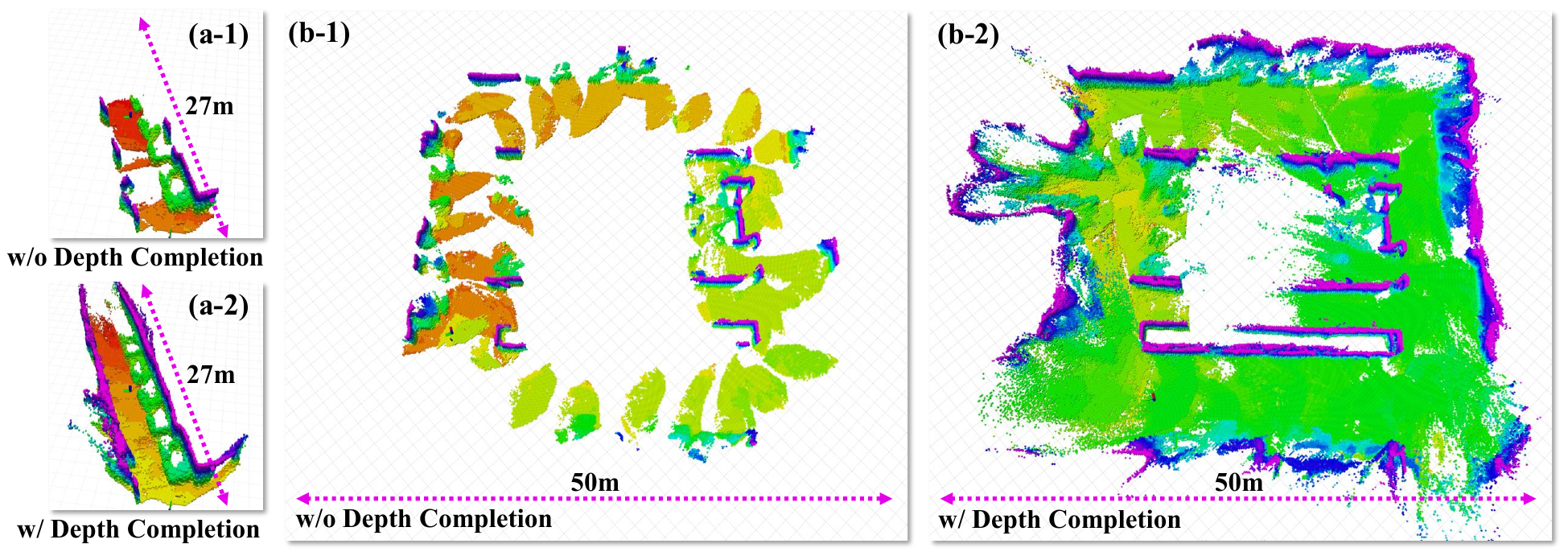}}
    \caption{\textbf{Large-scale grid mapping experiments.} (a) A long corridor. (b) A circular large-scale environment.}
    \label{real_world2}
\end{figure*}

\section{Robot Application Experiments}
\subsection{Edge-computing Applicability and Experimental Setup}

\textbf{Edge-computing Applicability.} To evaluate the performance of our proposed method in downstream robotic tasks, we first validate the edge-computing applicability for depth completion models. Table~\ref{cost} reports a comparison of computational costs. Except for the last row, all methods are evaluated for inference at 320×240 resolution on a single RTX 3090 GPU. Our method ranks second in both parameter count and FLOPs (second only to MDANet~\cite{MDANet}), and ranks second in average inference runtime (second only to NLSPN~\cite{NLSPN}), while achieving best depth completion accuracy. We further compute the average computational cost of selected methods except OMNI-DC~\cite{zuo2025omni}. Compared to this average baseline, ToFormer reduces the number of parameters, FLOPs, and runtime by 85.9\%, 48.6\%, and 73.8\%, respectively, demonstrating its superior efficiency. Then, we directly deploy our model to a Jetson Orin NX using libTorch without additional model pruning or quantization, obtaining an average runtime of approximately 10\,Hz for real-time tasks. 

\begin{figure}[t]
    \captionsetup{font={small}}
    \centerline{\includegraphics[width=0.42\textwidth]{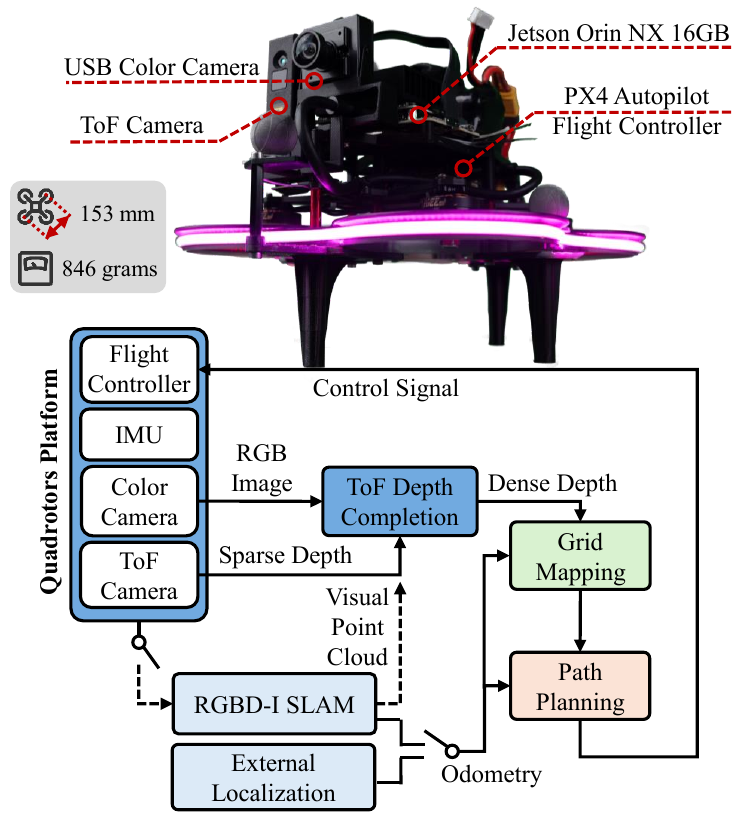}}
    \caption{Illustration of our quadrotor’s hardware and onboard algorithm with ToF depth completion model integrated.}
    \label{real_world1}
\end{figure}

\textbf{Experimental Setup.} As shown in Fig.~\ref{real_world1}, we build a PX4-based quadrotor platform equipped with a USB camera, a ToF camera, and Jetson Orin NX 16GB to support algorithm deployment and robot experiments. Specifically for onboard algorithm, the quadrotor can either run RGBD-I SLAM to provide visual point clouds and odometry, or rely on external localization (e.g., real-time kinematic~\cite{moon2016outdoor}, motion capture system~\cite{nokov}, or relative localization~\cite{li2025crepes}) to obtain only odometry. Our depth completion model takes RGB images from the color camera and sparse depth maps from the ToF camera as input, with the option to additionally use visual point clouds from SLAM to enhance completion performance. The mapping node utilizes dense depth maps generated by the depth completion node along with odometry to construct a dense probabilistic grid map. Based on the grid map and odometry, the path planning node performs motion planning and sends control commands to the flight controller.

\begin{figure*}[th]
    \captionsetup{font={small}}
    \centerline{\includegraphics[width=1.0\textwidth]{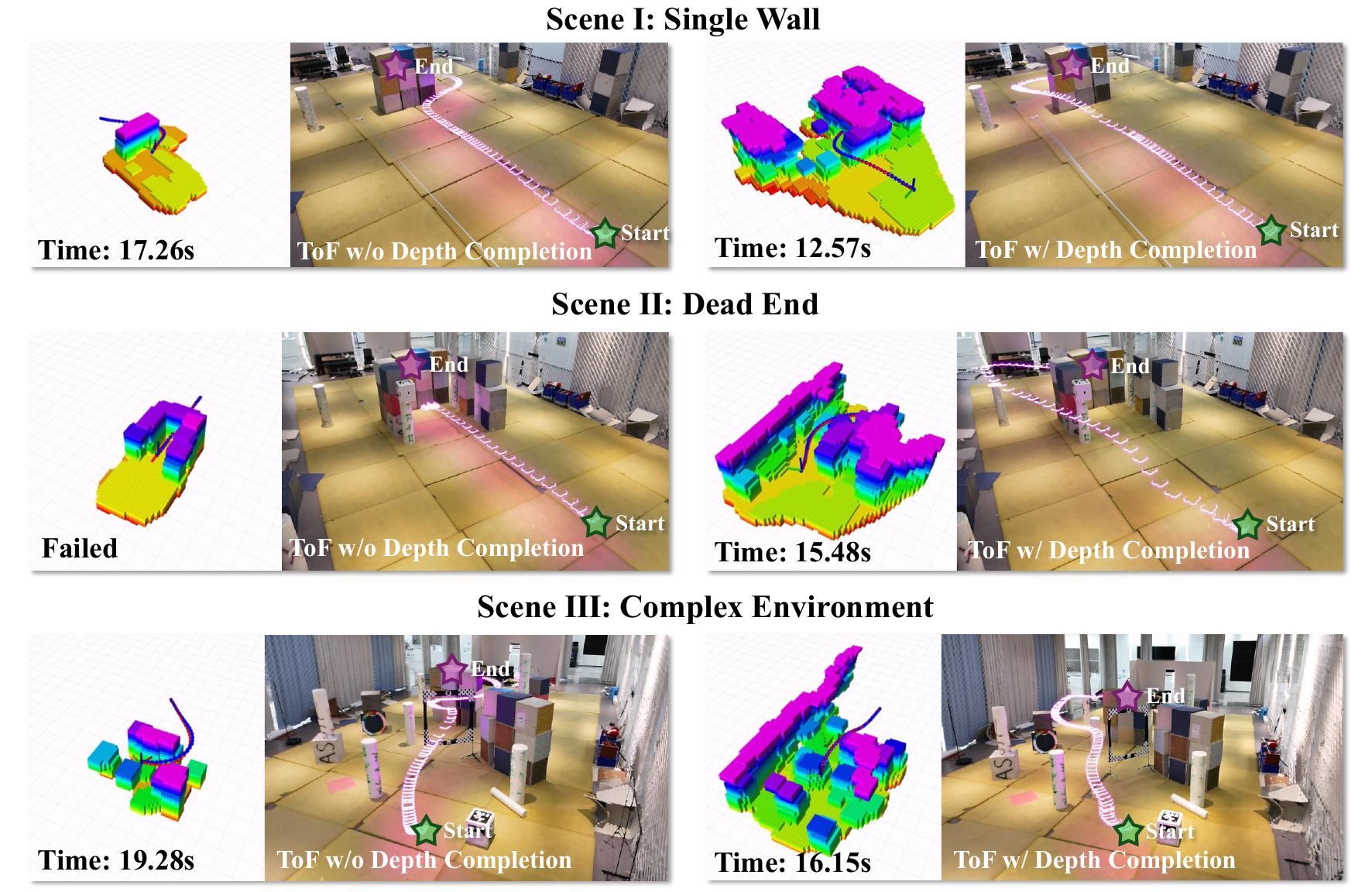}}
    \caption{\textbf{Long-range path planning experiments across three environments with increasing complexity.}}
    \label{real_world3}

\end{figure*}

Next, based on above preparation, we conducted grid mapping experiments (Fig.~\ref{real_world2}) and quadrotor path planning experiments (Fig.~\ref{real_world3}) in several large-scale environments.

\subsection{Large-Scale Grid Mapping Experiment}
In the grid mapping experiment, RGB images, ToF depth maps, and IMU data from the PX4 flight controller are subscribed by ORB-SLAM3 \cite{ORB-SLAM3}. We extracted visual point clouds from the local map of ORB-SLAM3 and fused them with the raw ToF depth to obtain ToF\&Visual sparse depth maps. Then, the depth completion node generates better dense depth maps. The dense depth maps and odometry information are fed into the grid mapping node to achieve dense reconstruction. Here, we set the maximum distance of ray casting in the mapping nodes to 15\,m to avoid excessive computational overhead.

Fig.~\ref{real_world2} (a-1) and (a-2) show the mapping results in a long corridor when using raw ToF depth and completed depth, respectively. When the quadrotor moves forward to the same position, our proposed method can directly reconstruct structures up to 15\,m ahead (upper limit of ray casting), whereas the raw ToF depth can only perceive structures within about 3\,m. Moreover, the reconstructed corridor based on completed depth exhibits a more complete geometric structure, effectively avoiding the large map holes caused by using short-range raw ToF depth.

Fig.~\ref{real_world2} (b-1) and (b-2) provide a more comprehensive evaluation in a  large-scale circular environment (50m$\times$50m). Without depth completion, only partial ground and wall regions are perceived, with severe structural incompleteness. In contrast, with depth completion, both ground and wall surfaces are reconstructed with high completeness, and the reconstructed walls maintained good planarity. These experiments demonstrate that the proposed method has the potential to extend the applicability of ToF cameras from small enclosed spaces to large-scale environments.

\begin{table}[th]
    \centering
    \captionsetup{font={small}}
    \caption{Quantitative results of the path planning experiment across three scenes.}
    \resizebox{0.476\textwidth}{!}{
    \begin{tabular}{c|cccc}
    \toprule
    \multicolumn{5}{c}{\textbf{Scene \uppercase\expandafter{\romannumeral 1}: Single Wall}} \\
    \midrule
    ToF Depth & \makecell[c]{Energy Cost$\downarrow$\\($m^{2} / s^{5}$)} & \makecell[c]{Traj. Length$\downarrow$\\($m$)}  & \makecell[c]{Time$\downarrow$\\($s$)}  & \makecell[c]{Avg. Speed$\uparrow$\\($m/s$)}  \\
    \midrule
    w/o DC  & 7.87  & 13.29 & 17.26 & 0.77\\
    w/ DC & 5.93  & 10.18 & 12.57 & 0.81\\
    \midrule
    \multicolumn{5}{c}{\textbf{Scene \uppercase\expandafter{\romannumeral 2}: Dead End}} \\
    \midrule
    w/o DC  & - & - & - & - \\
    w/ DC & 7.51  & 13.00   & 15.48 & 0.84 \\
    \midrule
    \multicolumn{5}{c}{\textbf{Scene \uppercase\expandafter{\romannumeral 3}: Complex Environment}} \\
    \midrule
    w/o DC  & 12.29  & 14.46 & 19.28 & 0.75 \\
    w/ DC & 8.73   & 13.24 & 16.15 & 0.82 \\
    \bottomrule
    \end{tabular}
    }
    \label{real_world_data}
\end{table}

\subsection{Path Planning Experiment}
In the path planning experiment, we adopt the classical Ego-Planner~\cite{zhou2020ego} for motion planning and evaluate the quadrotor’s autonomous planning performance with or without depth completion. All parameters are kept consistent with the original Ego-Planner paper, except that the maximum planning horizon is increased to 15 m. Especially, we assume that SLAM faced severe degradation or complete failure. Therefore, the depth completion model takes only the RGB image and the raw ToF depth map as input, while the quadrotor receives external localization information from a NOKOV motion capture system~\cite{nokov} as odometry. This setup directly challenges our method’s ability to infer long-range and large missing depth regions from limited near-range depth observations. 

Fig.~\ref{real_world3} visualizes the path planning performance of the quadrotor with or without depth completion across three environments of increasing complexity. And Table~\ref{real_world_data} presents the quantitative results of this experiment.
\begin{itemize}
    \item In Scene \uppercase\expandafter{\romannumeral 1}, a distant wall serves as an obstacle. Due to the limited sensing range of the raw ToF depth, the quadrotor can only detect the wall after approaching it, resulting in delayed path adjustments. With depth completion enabled, the quadrotor is able to detect the obstacle in advance, which results in reductions of 24.7\%, 23.4\%, and 27.2\% in energy cost, trajectory length, and travel time, respectively, along with a 5.2\% increase in average speed.
    \item In Scene \uppercase\expandafter{\romannumeral 2}, a dead-end corridor brings more challenges. The quadrotor relying on raw ToF depth fails to anticipate the dead end and ultimately become trapped, resulting in a planning failure. In contrast, with depth completion, the quadrotor successfully recognizes the dead end earlier and intelligently bypasses it.
    \item In Scene \uppercase\expandafter{\romannumeral 3}, we build a cluttered environment. The quadrotor without depth completion struggles to avoid obstacles that only become visible at close range, leading to frequent detours and slowdowns. With depth completion enabled, the quadrotor detects a spacious area on the left side of the scene and generates a more efficient path. As a result, the energy cost, trajectory length, and travel time are reduced by 29.0\%, 8.4\%, and 16.2\%, respectively, while the average speed increases by 9.3\%.
\end{itemize}

These experiments demonstrate that our method can be effectively integrated into time-critical path planning tasks, mitigating the local optimality issues caused by limited sensing range and significantly improving planning efficiency.

\section{Conclusion}\label{sec:conclusion} 

In this paper, we present ToFormer, an innovative framework designed to 
overcome the range limitation of ToF cameras. By developing a multi-sensor platform and a reconstruction-based data collection method, we create LASER-ToF, the first dataset and benchmark for large-scale ToF depth completion. By focusing on non-uniform nature of ToF depth map, we propose a novel depth completion network architecture, which captures long-range relationships for depth-missing regions, aggregates non-local point cloud geometry, and performs efficient 3D-2D fusion. Owing to these designs, ToFormer expands the sensing range of lightweight ToF cameras with advanced accuracy. Robot application experiments demonstrate the potential of our proposed method in downstream robot tasks. For future work, we will focus on tighter collaboration with SLAM systems and explore further applications in diverse robot task scenarios.

\bibliographystyle{IEEEtran}
\bibliography{main}

\end{document}